\documentclass{article}

\newcommand{\cm}{\textit{CroissantMiner}\xspace}

\usepackage[preprint, eandd]{neurips_2026}

\usepackage[utf8]{inputenc}
\usepackage[T1]{fontenc}
\usepackage{hyperref}
\hypersetup{
  pdfauthor={Berke Arda and Ahmetcan Yavuz and Paul Gerry and Sebastian Lobentanzer and Nobin Sarwar and Joan Giner-Miguelez and Kongtao Chen and Luyao Zhang and Mrinmaya Sachan and Mubashara Akhtar},
  pdftitle={CroissantMiner: Automated Extraction and Validation of Croissant Metadata for ML Datasets},
  pdfkeywords={dataset documentation, metadata extraction, large language models, responsible AI, Croissant, benchmark, LLM-as-judge, structured prediction}
}
\usepackage{url}
\newcommand{\linkicon}[1]{\raisebox{-0.18em}{\includegraphics[height=1.0em]{figures/icons/#1.pdf}}}
\usepackage{booktabs}
\usepackage{amsfonts}
\usepackage{nicefrac}
\usepackage{microtype}
\usepackage{xcolor}
\usepackage{graphicx}
\usepackage{multirow}
\usepackage{listings}
\usepackage{tikz}
\usepackage{xspace}
\usepackage{enumitem}
\usepackage{placeins}
\usepackage{makecell}
\usepackage{wrapfig} 
\usepackage[most]{tcolorbox}
\usepackage{comment}

\usetikzlibrary{positioning , arrows.meta, calc}

\title{\raisebox{-0.3\height}{\includegraphics[height=2.4em]{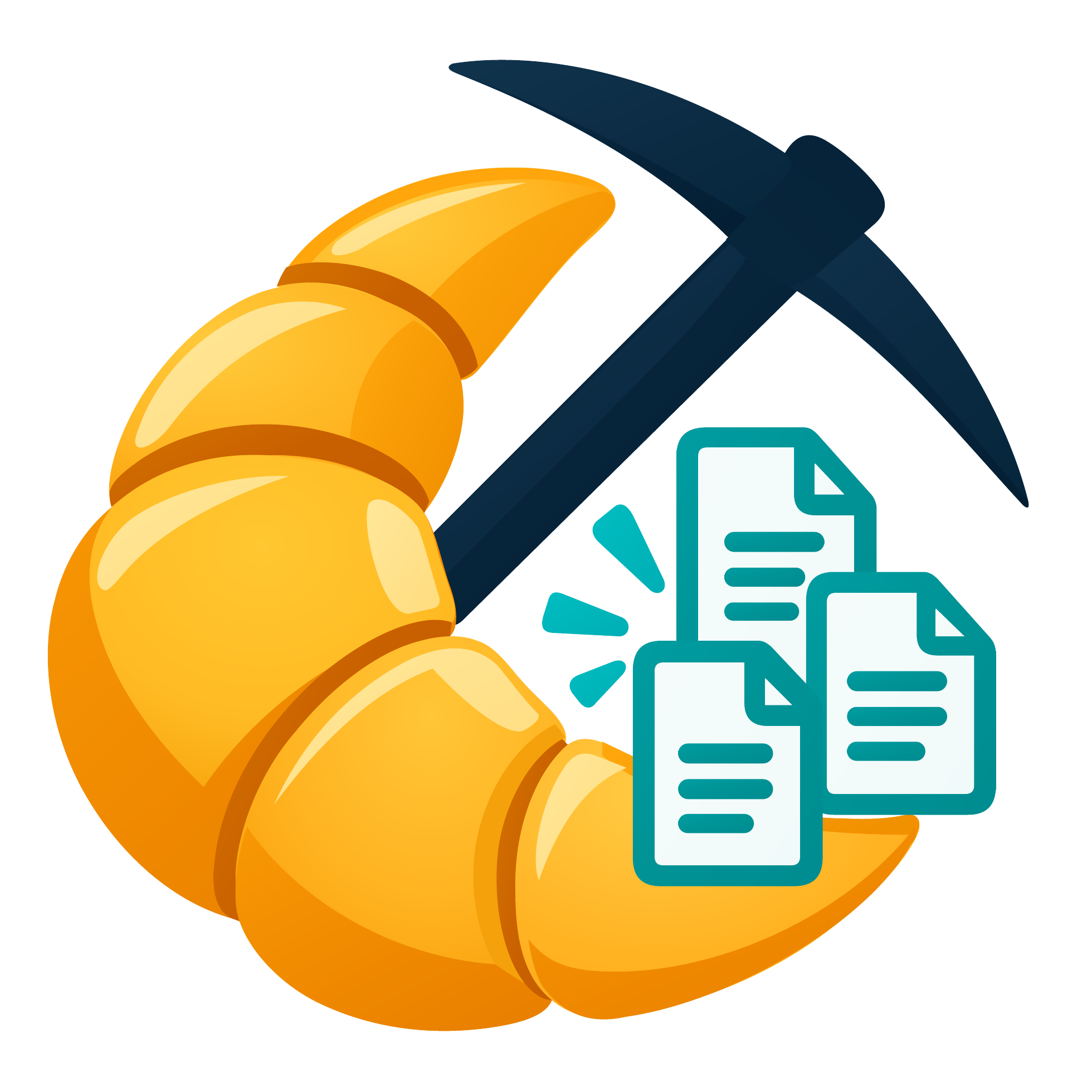}}\hspace{0.4em}CroissantMiner: Automated Extraction and Validation of Croissant Metadata for ML Datasets}

\author{%
  \textbf{Berke Arda}\textsuperscript{*1} \quad
  \textbf{Ahmetcan Yavuz}\textsuperscript{*1} \quad
  \textbf{Paul Gerry}\textsuperscript{1,3} \quad
  \textbf{Sebastian Lobentanzer}\textsuperscript{4} \quad \\[2pt]
  \textbf{Nobin Sarwar}\textsuperscript{5} \quad
  \textbf{Joan Giner-Miguelez}\textsuperscript{6} \quad
  \textbf{Kongtao Chen}\textsuperscript{7} \quad
  \textbf{Luyao Zhang}\textsuperscript{8} \quad \\[2pt]
  \textbf{Mrinmaya Sachan}\textsuperscript{1,2} \quad
  \textbf{Mubashara Akhtar}\textsuperscript{1,2} \\[6pt]
  \normalfont\small
  \textsuperscript{*}Lead authors \quad \\[2pt]
  \textsuperscript{1}ETH Zurich \quad
  \textsuperscript{2}ETH AI Center \quad
  \textsuperscript{3}CSAIL, MIT \quad
  \textsuperscript{4}Helmholtz Zentrum M\"unchen \\[2pt]
  \textsuperscript{5}University of Maryland, Baltimore County \quad
  \textsuperscript{6}Barcelona Supercomputing Center \\[2pt]
  \textsuperscript{7}Google \quad
  \textsuperscript{8}Duke Kunshan University
}

\begin{document}

\maketitle

\begin{abstract}
Croissant has emerged as a standard for machine-readable dataset metadata, yet populating its fields remains labor-intensive and requires careful reading of accompanying dataset documentation.
We present the first benchmark enabling end-to-end evaluation of metadata extraction aligned with a community-standard schema.
The benchmark comprises 602 papers, including 102 with human-validated gold annotations and 500 with LLM-generated silver annotations, covering the full Croissant schema with both core and Responsible AI (RAI) fields.
Using this benchmark, we evaluate a range of extraction systems spanning frontier models, open-weight models, and agentic architectures, under a two-tier evaluation framework that combines rule-based scoring with an LLM judge selected via human audit.
We find that single-pass extraction consistently outperforms the four agentic architectures we evaluate: across backbones, these decomposed variants achieve lower accuracy than a single full-context pass. The largest gap appears on long-form RAI fields, which require synthesizing and interpreting information scattered across a paper rather than copying it from a single location, a setting where current systems remain far from reliable.
We release the benchmark, evaluation code, judge audit, a live demo, and a leaderboard open to new systems.

\vspace{0.3em}
\begin{center}\hypersetup{pdfborder={0 0 0}}
\href{https://berkearda.github.io/croissantminer/}{\linkicon{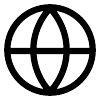}\ berkearda.github.io/croissantminer}
\end{center}

\end{abstract}

\section{Introduction}
\label{sec:intro}

Machine Learning (ML) research relies on well-documented datasets that are discoverable, interpretable, and reproducible~\citep{luccioni2022framework, bhardwaj2024machine, akhtar2024croissant, seidenberger2025big, winata2025datasheets, jia2025sustainable, reynolds2026reflection}. In practice, however, dataset documentation remains fragmented, inconsistent, and difficult to integrate across platforms, motivating machine-readable metadata standards. Differences in formats, limited interoperability across tools, and sparse descriptions of data collection and usage hinder dataset discovery, reuse, and reproducibility, while also raising concerns around licensing, provenance, and responsible use~\citep{dodge2021documenting, peng2021mitigating, longpre2024large, zhao2024taxonomy, liu2024best, kim2025not}.

\begin{figure}[!htbp]
  \centering
  \includegraphics[width=0.98\linewidth]{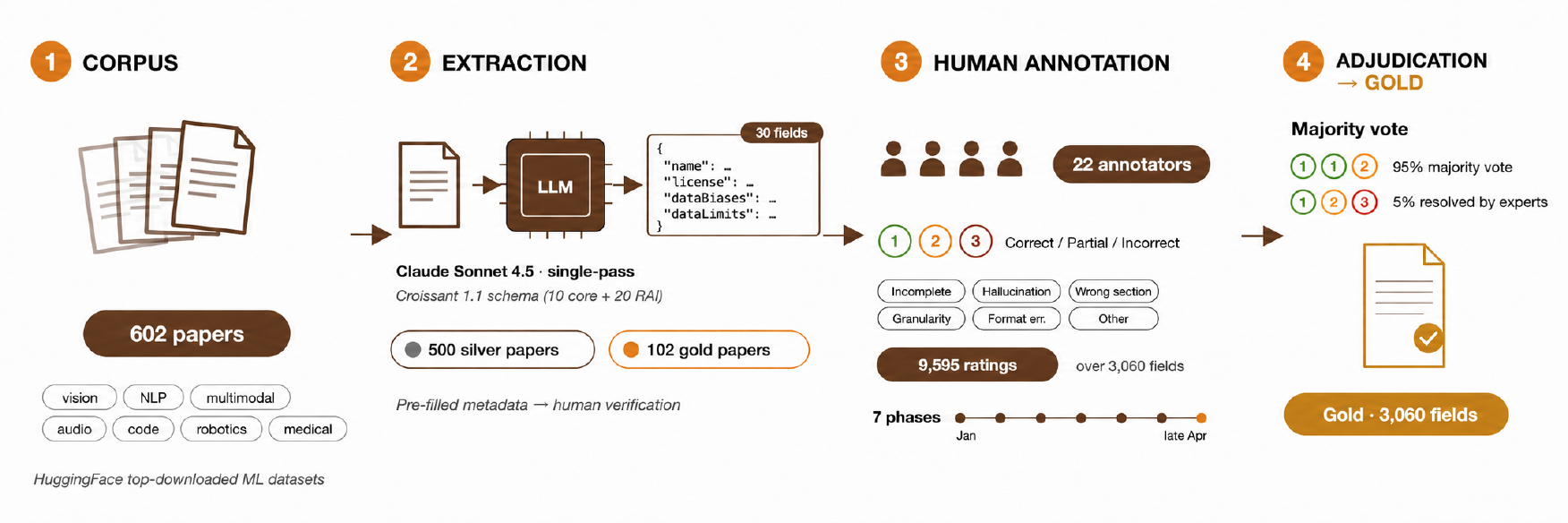}
  \vspace{-5mm}
  \caption{\textbf{The CroissantMiner data-creation pipeline.}
    \textbf{(1) Corpus:} ML dataset papers from top-downloaded Hugging Face datasets (vision, NLP, audio, etc.).
    \textbf{(2) Extraction:} a single-pass LLM generates 30-field Croissant metadata per paper, forming the silver split and the pre-fills for the gold split.
    \textbf{(3) Annotation:} annotators rate each pre-fill against the source paper on a 3-level rubric (Correct / Partially Correct / Not Correct) with failure-mode labels (e.g., hallucination, incomplete), producing 9{,}595 ratings over 3{,}060 cells.
    \textbf{(4) Adjudication:} majority vote resolves 95\% of cells; senior-author review resolves the rest.}
  \label{fig:croissantminer_overview}
  \vspace{-10pt}
\end{figure}

The Croissant metadata format~\citep{akhtar2024croissant}, recently adopted as a community standard, addresses these challenges by providing a structured, machine-readable schema for describing ML datasets' attributes, resources, and responsible AI properties. Major platforms including Hugging Face, Kaggle, and OpenML now support Croissant, and NeurIPS 2026 mandates that all dataset submissions include Croissant Responsible AI (RAI) metadata files~\citep{neurips2026edfaq}. Post-acceptance surveys from the NeurIPS 2025 Datasets and Benchmarks track report that 16\% of authors encountered submission difficulties, primarily due to dataset hosting constraints and Croissant metadata generation challenges~\citep{neurips2025db}. Despite this growing adoption, creating Croissant metadata remains a manual and time-consuming process that requires careful reading of dataset papers and accompanying documentation. With over 500{,}000 public datasets on Hugging Face and more than 32{,}000 datasets created monthly in 2025~\citep{hfhubstats2026, hf_state_of_os_spring26}, manual metadata creation cannot scale. Existing automated approaches remain limited, typically focusing on a small subset of fields and struggling with the variability and complexity of scientific papers~\citep{alyafeai2025mole, liu2024cardgen}. 

We introduce \textbf{\cm}, a benchmark for automated extraction of Croissant metadata from ML dataset papers. It pairs 602 dataset papers, 102 with human-validated gold annotations and 500 with LLM-generated silver annotations, with the full 30-field Croissant~1.1 schema, covering both core fields and all Responsible-AI attributes. To our knowledge, this is the first benchmark enabling end-to-end evaluation of metadata extraction aligned with a community-standard schema on real dataset papers.

We evaluate 24 extraction systems, including frontier and open-weight models as well as four agentic architectures that decompose the task into multiple steps. Among the systems we evaluate, single-pass extraction with Claude Sonnet~4.6 and a canonical prompt scores highest. In contrast, the four agentic architectures consistently underperform single-pass extraction with the same backbone, suggesting that task decomposition can degrade performance for schema-complete metadata extraction. This gap is most pronounced on long-form RAI fields, where accurate extraction requires integrating information across multiple sections of the paper. We also release an open-source system and demo that generate metadata drafts for authors, reviewers, and users to inspect and refine. Our contributions are as follows:

\vspace{-5pt}

\begin{enumerate}[leftmargin=25pt, itemsep=2pt, parsep=0pt]
    \item \textbf{A benchmark for Croissant metadata extraction}, with human-validated gold annotations from 22 annotators providing 9{,}595 ratings and LLM-generated silver annotations, covering the full 30-field Croissant~1.1 schema including all Responsible-AI fields.
    \item \textbf{A systematic evaluation enabled by this benchmark}, comparing frontier models, open-weight models, and four agentic architectures under a two-tier scoring framework that combines rule-based metrics with a human-audited LLM judge.
    \item \textbf{Findings on extraction difficulty}, showing that single-pass extraction consistently outperforms the four agentic architectures we evaluate, that long-form RAI fields requiring synthesis across a paper are the hardest to extract, and an analysis of annotation reliability and field-level difficulty.
    \item \textbf{Public resources}, released as open-source software and a Hugging Face Space for generating Croissant metadata from dataset papers for authors, reviewers, and users.
\end{enumerate}

\section{Related Work}

\paragraph{From Documentation Frameworks to Machine-Readable Standards}
The recognition that ML datasets require systematic documentation has produced a broad landscape of frameworks. Prior work introduced Datasheets for Datasets~\citep{gebru2021datasheets}, Data Statements~\citep{bender2018data}, Model Cards~\citep{mitchell2019model}, Data Cards~\citep{pushkarna2022data}, and Data Nutrition Labels~\citep{holland2018dataset}. However, these frameworks largely produce natural-language documentation that is costly to create, difficult to standardize, and hard to reuse at scale. Croissant~\citep{akhtar2024croissant} addresses this limitation by providing a machine-readable metadata standard for ML datasets, including a Responsible-AI (RAI) extension. Despite this adoption, a key gap remains: while platforms can infer structural metadata from data files, contextual and RAI information remains embedded in dataset documentation. The \cm benchmark targets this bottleneck by evaluating the extraction of these fields directly from scientific documents.

\paragraph{Automated Metadata and Documentation Generation}
Several systems address the challenge of automating dataset and model documentation. CardGen~\citep{liu2024cardgen} generates free-text cards from papers and repositories, while Datadoc Analyzer~\citep{giner2023datadoc} extracts structured attributes for data papers. MOLE and MeXtract~\citep{alyafeai2025mole, alyafeai2025mextract} extract metadata attributes using LLMs, but target alternative schemas (e.g., Masader) and do not produce Croissant-conformant output. Platform-based tools (e.g., Hugging Face) automatically generate metadata from dataset files, but are limited to structural fields. These approaches are complementary to the \cm benchmark, which focuses on the extraction of contextual and RAI information from dataset papers. Datasets for benchmarking metadata extraction remain limited and schema-specific~\citep{akhtar2024croissant}.

\section{The CroissantMiner Benchmark}
\label{sec:annotation}

We release the \cm benchmark for end-to-end, schema-complete metadata extraction from ML dataset papers. The dataset pairs 602 ML dataset papers (102 human-validated gold and 500 silver) with the full 30-field Croissant 1.1 schema, enabling evaluation of metadata extraction on real dataset papers where information is often ambiguous, scattered across sections, or missing~\citep{dodge2021documenting, longpre2024large}.

\textbf{Paper Selection and Benchmark Construction} We construct the benchmark from publicly available AI dataset papers, prioritizing highly-downloaded datasets on Hugging Face. The gold split was manually curated across diverse modalities and formats, while the silver split was selected automatically from the Hugging Face Hub (Section~\ref{ssec:silver}). The corpus spans vision, NLP, multimodal learning, audio, code, robotics, and medical AI.

\textbf{Metadata Schema and Annotation Targets} Each paper is annotated against the Croissant 1.1 schema,\footnote{\url{https://docs.mlcommons.org/croissant/docs/croissant-spec-1.1.html}} yielding 30 target fields: 10 core descriptive (\texttt{sc:}/\texttt{cr:}) and 20 Responsible AI (\texttt{rai:}) fields (see Appendix~\ref{app:fields} for the full field list with definitions and coverage). These fields range from constrained categorical values (e.g., \texttt{license}, \texttt{inLanguage}) to short-text (e.g., \texttt{creator}) to long-form prose (e.g., \texttt{rai:dataCollection}). Many RAI fields are frequently not reported in dataset papers, so correctly leaving a field empty when the paper does not report it is as important as extracting it accurately. Gold values record only what the paper itself states: a field the paper does not document is marked \texttt{[NULL - not found in paper]}, even when the information exists elsewhere, such as on a dataset card.

\subsection{Gold Dataset: Human Validation Protocol}
\label{ssec:gold}

We adopt a pre-fill paradigm: an LLM first generates a complete metadata record for each paper, which annotators then verify and correct rather than extract from scratch (see Fig.~\ref{fig:croissantminer_overview}). This design reduces annotation effort and improves consistency across fields. We recruited 22 annotators with experience in ML evaluation, benchmarking, and ontology engineering, spanning 18 affiliations across academia, industry, and research institutes; details in Appendix~\ref{app:annotation-phases}. They joined through an open research call and participated voluntarily without financial compensation; eight are co-authors of this paper.

\textbf{Collection Process.} The LLM-based pre-fill extraction of metadata was performed using Claude Sonnet 4.5 over all 602 papers, populating the 30-field schema at temperature 0 with a canonical prompt (Appendix~\ref{app:prompts}). Claude Sonnet 4.5 was selected after pilot extractions on a 10-paper subset against GPT-4o-mini and Gemini 2.5 Pro, where it produced the most schema-conformant JSON-LD output and the fewest hallucinated fields. Annotators were shown the dataset identifier, paper URL, field name, and pre-filled value, and were asked to assess the pre-filled values against the source paper instead of re-extracting metadata from scratch. Each paper-field pair received at least three independent ratings on a three-level rubric (\textit{Correct} / \textit{Partially Correct} / \textit{Not Correct}), as well as failure-mode labels (incomplete, hallucination, wrong section, granularity mismatch, format error, other), free-text notes, and confidence scores. In total, we collected 9{,}595 ratings for 3{,}060 metadata fields across 102 papers. Majority voting resolved 95.0\% of cells; a senior author adjudicated 140 of the remaining cells, and the lead author resolved the other 14. Final gold labels were derived via per-field majority vote.\footnote{Appendix~\ref{app:annotation-phases} further details the data collection and annotation process.} Of all ratings, 83.4\% judged the pre-filled value fully correct and 16.6\% flagged it as partially correct or not correct; annotators supplied an explicit correction in 12.1\% of ratings, at similar rates for core and RAI fields, 11.9\% and 12.2\%, so the gold labels reflect substantial human review rather than the pre-fill alone.

\textbf{Annotator Selection and Validation.} Annotators were calibrated through a pilot annotation round in which annotators rated the same shared sample of pre-filled cells; the authors then evaluated submissions and checked for rubric adherence, quality of corrections, and accuracy with respect to the source papers. Based on this assessment, annotators were grouped into quality tiers and assigned tasks accordingly, with higher-tier annotators handling more complex metadata fields (e.g.\ RAI fields). Workload was also balanced based on self-reported availability.
During the follow-up annotation phase, submissions were continuously spot-checked, and assignments were adjusted when ratings did not align with the source papers. To ensure coverage and consistency, we tracked annotator participation and redistributed tasks as needed. Additional details on annotator selection, assignment, and validation are provided in Appendix~\ref{app:annotation-phases}.

\textbf{Inter-annotator Agreement.} Agreement is high overall: 61.5\% of cells are unanimous. To quantify this, we report Gwet's AC$_1$ in addition to Krippendorff's $\alpha$, since when ratings cluster on \emph{Correct}, $\alpha$ can underestimate agreement despite high consistency. Agreement nonetheless varies substantially across fields: constrained fields achieve near-perfect agreement (e.g., \texttt{cr:isLiveDataset}, AC$_1 = 0.96$), while open-ended RAI fields show the lowest scores (\texttt{rai:dataAnnotationAnalysis}, AC$_1 = 0.45$; \texttt{rai:dataPreprocessingProtocol}, AC$_1 = 0.46$). Low agreement is not caused by missing documentation: rarely documented fields reach the highest agreement (e.g., \texttt{rai:dataImputationProtocol}, AC$_1 = 0.93$), since annotators agree that the field is absent. Disagreement instead concentrates in fields that require interpretation. For \texttt{sc:publisher} (AC$_1 = -0.09$), annotators systematically disagree between dataset host, venue, and author affiliation, indicating ambiguity in the schema definition rather than annotator error.\footnote{A breakdown of annotation agreement per field and further analysis is provided in Appendix~\ref{app:annotation_reliability}.}

\subsection{Silver Dataset: Annotations at Scale}
\label{ssec:silver}

The silver split complements the gold benchmark by enabling model development (distillation, fine-tuning), large-scale corpus analysis, and metadata-trend studies that would otherwise be impractical on the manually validated subset.
We release a 500-paper silver split with AI-generated Croissant metadata. The 500 papers were selected from public Hugging Face datasets that link to an arXiv paper, ranked by download count, deduplicated by arXiv identifier, and filtered to remove overlap with the gold split. Its metadata is the Sonnet 4.5 pre-fill described in Section~\ref{ssec:gold}, without human review, so silver labels should not be treated as equivalent to gold (Figure~\ref{fig:silver_vs_gold} compares silver and gold per field).

\subsection{Dataset Statistics and Coverage}

Across all 602 papers, an average of 19.2 out of 30 fields (64.0\%) are documented, with higher coverage in the gold split (21.7) than in the silver split (18.7). Coverage also varies by domain: question-answering benchmarks document the most fields on average (21.0), while robotics datasets document the fewest (17.7). Coverage differs most between core and RAI fields. Core fields such as \texttt{name}, \texttt{description}, and \texttt{creator} appear in over 99\% of papers, while RAI fields show very uneven coverage. Some fields are rarely reported in dataset papers (e.g., \texttt{rai:dataImputationProtocol}, \texttt{rai:dataCollectionMissingData}, \texttt{rai:dataCollectionTimeframe}, all below 25\%), while others are commonly included (e.g., \texttt{rai:dataUseCases}, \texttt{rai:dataLimitations}, above 90\%). These coverage statistics quantify the difficulty of the benchmark. First, extraction systems must correctly identify absent information: in the gold split, 27.7\% of all cells have no documented value, and the rate is nearly three times higher for RAI fields (35.6\%) than for core fields (11.9\%). Leaving such fields empty instead of guessing is therefore a substantial part of the task. Second, the fields that are documented often require synthesizing information from multiple sections of a paper, and these long-form RAI fields show both the lowest annotator agreement and the largest system errors (Section~\ref{sec:experiments}). The full per-field breakdown is provided in Appendix~\ref{app:perfield} (Figure~\ref{fig:field_fill_rates} visualizes the rates).

\section{The \cm System}
\label{sec:system}
\label{sec:task-pipeline}

In addition to the benchmark, we introduce the \cm system for automatically extracting structured Croissant metadata from dataset papers. We compare a simple single-pass LLM-based approach against four agentic architectures that decompose the extraction task in different ways. We formulate extraction as a structured prediction task: given a dataset paper as PDF, the system returns a Croissant 1.1 JSON-LD record with all 30 fields (Appendix~\ref{app:fields}). The extraction pipeline has three stages: (1) \textbf{Text Extraction:} Convert the input PDF to plain text using PyPDF2 and remove the reference sections by regex-matching; appendices are retained. (2) \textbf{LLM Extraction:} Process the cleaned text using the canonical prompt, in a single call for the single-pass systems or through the architecture-specific procedures of Section~\ref{sec:agentic-variants}. (3) \textbf{Validation:} Parse the JSON output (handling occasional Markdown code-fence wrapping) and check it against the 30-field schema. Schema violations occurred in fewer than 1\% of extractions.

\begin{figure}[!t]
  \centering
  \includegraphics[width=\linewidth]{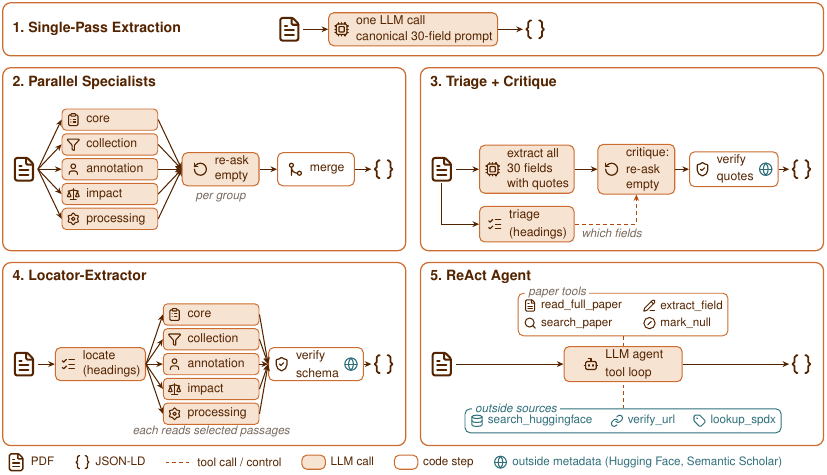}
  \caption{\textbf{The four agentic architectures evaluated against single-pass extraction.} All systems take a paper PDF as input and return a 30-field Croissant JSON-LD record (Section~\ref{sec:agentic-variants}). They differ in how the extraction is split into LLM calls and code steps, and three of them can also fill some fields from outside metadata.}
  \vspace{-3mm}
  \label{fig:architectures}
\end{figure}

\subsection{System Architectures}
\label{sec:agentic-variants}
\label{sec:prompt-design}

Figure~\ref{fig:architectures} shows the single-pass baseline and the four agentic architectures.\footnote{All discussed systems are released as an open-source Python package at \url{https://github.com/berkearda/croissantminer}, and a public demo is hosted on a Hugging Face Space at \url{https://huggingface.co/spaces/bearda/croissantminer}. Implementation details are given in Appendix~\ref{app:implementation}.} The agentic architectures received comparable prompt-engineering effort: each uses per-field extraction guidance analogous to the canonical prompt, their prompts were refined over multiple iterations (including anti-hallucination instructions and a verify-and-correct pass), and we report the best-performing prompt variant for each architecture. We select these four architectures to cover the main decomposition patterns in current agentic systems: role-based task decomposition (Parallel Specialists), plan-then-execute pipelines with self-critique (Triage~+~Critique), retrieval-based context narrowing (Locator-Extractor), and dynamic tool-use loops (ReAct)~\citep{yao2023react}. Each pattern represents a distinct hypothesis about how decomposition could help schema-complete extraction, and covering all four allows us to attribute performance differences to the decomposition strategy rather than to a single design choice.

\textbf{1. Single-Pass Extraction with Canonical Prompt} To assess whether state-of-the-art LLMs can perform schema-complete metadata extraction in a single pass over the full paper, we design a structured canonical prompt covering all 30 Croissant fields.
Our structured prompt combines four elements:
\textit{(i)} role framing (metadata extraction expert),
\textit{(ii)} accuracy-first constraints that forbid using external knowledge,
\textit{(iii)} per-field instructions for all 30 schema fields, and
\textit{(iv)} targeted guidance for fields prone to hallucination.
Each field guide includes a definition, positive and negative examples, and explicit instructions to return null when information is not present. To ensure determinism, we fix temperature=0, max\_tokens=4096, and the model version, and log the prompt hash for reproducibility. The full prompt is provided in Appendix~\ref{app:prompts}.

\textbf{2. Multi-Agent: Parallel Specialists} The \emph{Parallel Specialists} architecture splits the 30 fields across five specialist LLM calls. The extraction is done in three stages:
\textit{(1) Parallel extraction.} Five specialist LLM calls (each with a separate prompt and field subset) run in parallel over the full paper, partitioning the 30 fields into disjoint groups: core, collection, annotation, impact, processing. 
These groups are aligned to Croissant's natural RAI sub-topics and their typical alignment with sections in dataset papers.
\textit{(2) Correction.} Each specialist is asked once more for the fields it returned empty.
\textit{(3) Aggregation.} Afterwards, the outputs are merged via dictionary union; the disjoint partition guarantees that no key conflicts occur.
Unlike Triage~+~Critique and Locator-Extractor, this architecture receives no guidance about where in the paper each field is likely to be documented.

\textbf{3. Triage \& Critique} The \emph{Triage \& Critique} architecture surrounds a full-paper extraction with a planning step and two checking steps.
\textit{(1) Triage.} A lightweight LLM call (Gemini~2.5~Flash) reads the paper's section headings and estimates, for each of five field groups (core, collection, annotation, impact, processing), whether the paper is likely to document it.
\textit{(2) Extraction.} A single LLM call reads the full paper and extracts all 30 fields, each with a supporting quote; the prompt asks the model to leave a field empty unless the paper discusses it explicitly.
\textit{(3) Self-critique.} Empty fields that the triage marks as likely present, and all empty core fields, are requested again in a second full-paper call.
\textit{(4) Verification.} Long-form RAI values whose supporting quote cannot be found in the paper are removed, and a few empty core fields (license, URL, language, citation) are filled from Hugging Face and Semantic Scholar metadata.
Full prompts are provided in Appendix~\ref{app:prompts}.

\textbf{4. Locator-Extractor} The \emph{Locator-Extractor} architecture decomposes extraction into three stages, including a verification step. Like Parallel Specialists, it uses one LLM call per field group, but each call sees only the passages selected for its group rather than the full paper.
\textit{(1) Locate.} An LLM call first analyzes the paper's section headings to identify likely locations of relevant information for each field group (Gemini~2.5~Flash; for the Sonnet~4.6 variant, Sonnet~4.6 itself; for the hybrid, Gemini~3.1~Pro), with a keyword-matching fallback when no sections are identified.
\textit{(2) Extract.} Five LLM calls, one per field group, each producing outputs in a \texttt{\{value, evidence\_quote\}} format.
\textit{(3) Verify.} For fields that remain empty, the system performs cross-document enrichment using external sources (Hugging Face metadata and Semantic Scholar lookup), followed by schema validation and per-field confidence scoring.

\textbf{5. ReAct Agent} For the \emph{ReAct Agent}~\citep{yao2023react}, we model extraction as a dynamic, multi-step reasoning process with tool use. Unlike the previous architectures, which follow fixed pipelines, the ReAct agent dynamically decides which fields to extract, which tools to invoke, and when to terminate the process. The agent alternates between reasoning and tool calls over multiple steps (up to a default budget of 20 turns), gradually constructing the output. The cap acts as a safety limit rather than a tuned parameter: in practice the agent stops after 3.8 to 4.9 turns on average (Sonnet~4.6: 4.9, maximum 6; GPT-5.4: 4.9, maximum 7; Gemini~3.1~Pro: 3.8), and it reached the cap in only 2 of 264 runs, both with Gemini~3.1~Pro. The full paper text and system prompt are sent at every turn; for the Claude models they are served from a prompt cache. The agent has access to a set of tools for reading, retrieving, and validating information, including document access (\texttt{read\_full\_paper}), paragraph retrieval (\texttt{search\_paper}), metadata lookup (\texttt{search\_huggingface}), URL verification, and license normalization. Three of the agentic architectures (Triage~+~Critique, Locator-Extractor, and ReAct) can also fill some fields from Hugging Face or Semantic Scholar metadata; single-pass extraction and Parallel Specialists use only the paper.

\section{Results \& Discussion}
\label{sec:experiments}

We evaluate a range of extraction systems on the \cm benchmark, including single-pass LLM extractors and the four agentic architectures described above. The gold split is divided into a 14-paper development set and an 88-paper held-out test set, and on the test split we analyze overall performance, field-level differences, architectural trade-offs, and associated failure and cost patterns.

\subsection{Experimental Setup}
\label{sec:experimental-setup}

\textbf{Backbone Models.} We evaluate seven proprietary frontier models (Claude Sonnet~4.5, Sonnet~4.6, Opus~4.7, GPT-5.4 full and mini, Gemini~3.1~Pro Preview, Gemini~2.5~Flash) and five open-weight models (Llama~4 Scout~17B, Qwen~3.6~35B-A3B and Mistral Small~4, served via vLLM on an institutional GPU cluster; DeepSeek~V3.2 and GLM-5.1, through a hosted API). Claude Sonnet~4.5 generated the gold pre-fills (Section~\ref{sec:annotation}), so it is reported as an unranked diagnostic in Table~\ref{tab:test88-headline} rather than ranked with the other systems.

\begin{wraptable}{r}{0.42\linewidth}
    \vspace{-12pt}
    \centering
    \scriptsize
    \setlength{\tabcolsep}{3.5pt}
    \vspace{-5pt}
    \caption{Judge selection: Cohen's quadratic-weighted $\kappa$ against human consensus on 60 development cells (30 calibration, 30 validation).}
    \vspace{2pt}
    \label{tab:judge-selection}
    \begin{tabular}{lccc}
    \toprule
    Candidate judge  & cal. & val. & pooled \\
    \midrule
    \textbf{GLM-5}   & 0.900 & 0.880 & \textbf{0.890} \\
    DeepSeek V3.2    & 0.828 & 0.882 & 0.855 \\
    GPT-5.5          & 0.784 & 0.904 & 0.847 \\
    Gemini 2.5 Pro$^\dagger$ & 0.900 & 0.743 & 0.823 \\
    Qwen 3 Max       & 0.784 & 0.857 & 0.823 \\
    Llama 4 Maverick & 0.781 & 0.855 & 0.821 \\
    \bottomrule
    \multicolumn{4}{l}{\tiny $^\dagger$Validation over 29 cells (one response could not be parsed).}
    \end{tabular}
    \vspace{-10pt}
\end{wraptable}
\textbf{Evaluation Metrics.} We use a two-tier evaluation pipeline: rule-based scoring for structured fields and an LLM judge on a 1--3 ordinal scale for long-form fields. Null predictions are handled explicitly (correct-null skipped; false positives penalized). Composite scores average over all 30 fields and are reported with 2{,}000-replicate paper-clustered bootstrap confidence intervals~\citep{efron1979bootstrap}. Pairwise comparisons use the paired bootstrap, Wilcoxon signed-rank~\citep{wilcoxon1945individual}, and McNemar's test~\citep{mcnemar1947note} with BH-FDR correction~\citep{benjamini1995controlling}.

\textbf{Judge Selection and Agreement.} We chose the Tier-2 judge by its agreement with human ratings, measured with Cohen's quadratic-weighted $\kappa$. Six candidate models rated 60 development cells, split into 30 calibration and 30 validation cells so that the choice does not rest on one small sample. GLM-5 agreed best with the human consensus ($\kappa = 0.890$) and was stable across both halves (0.900 and 0.880), while Gemini 2.5 Pro fell from 0.900 to 0.743 on validation (Table~\ref{tab:judge-selection}). These values were measured with the initial rubric and the gold labels before our audit; with the final rubric and the audited gold, GLM-5 reaches $\kappa = 0.708$ on the same 60 cells. To test the judge on harder cases, two authors independently rated 200 further cells covering all 20 RAI fields, with extra difficult cases, and a third author resolved their disagreements. The judge matches this human consensus in 71.5\% of cells ($\kappa = 0.663$), almost as often as the two raters match each other (72.0\%, $\kappa = 0.739$); when it disagrees, it is usually more lenient than the humans (45 of 57 cases). The judge is therefore a reasonable stand-in for human rating at this scale, but not a replacement for it (Section~\ref{sec:limitations}). Full details are in Appendix~\ref{app:prompts-judge}.

\begin{table}[!t]
    \centering
    \scriptsize
    \caption{\textbf{Results on the test split}, scored with the GLM-5 Tier-2 judge on audited gold samples (\S\ref{sec:experimental-setup}). \emph{Core} averages the 10 core fields (rule-based), \emph{RAI} the 20 RAI fields (LLM judge), and \emph{Composite} weights all 30 fields equally (2{,}000-replicate bootstrap CIs). Ranks are reported per architecture type. Anthropic-family systems are marked~$^{*}$. Claude Sonnet~4.5, which generated the gold pre-fills, is listed as an unranked diagnostic: its score largely reflects agreement with its own output and is not comparable with the other rows.}
    \label{tab:test88-headline}
    \resizebox{\textwidth}{!}{%
    \begin{tabular}{llcccc}
    \toprule
    Rank & System & Architecture & Core & RAI & Composite [95\% CI] \\
    \midrule
    1 & \textbf{Claude Sonnet 4.6}$^{*}$ & Single-Pass & \textbf{0.752} & 0.687 & \textbf{0.709}\, [0.688, 0.729] \\
    2 & Claude Opus 4.7$^{*}$ & Single-Pass & 0.676 & \textbf{0.711} & 0.699\, [0.665, 0.732] \\
    3 & GPT-5.4 & Single-Pass & 0.653 & 0.671 & 0.665\, [0.648, 0.692] \\
    4 & Qwen 3.6 35B-A3B & Single-Pass & 0.698 & 0.601 & 0.634\, [0.615, 0.654] \\
    5 & GLM-5.1 & Single-Pass & 0.675 & 0.599 & 0.625\, [0.604, 0.645] \\
    6 & Gemini 2.5 Flash & Single-Pass & 0.615 & 0.616 & 0.616\, [0.592, 0.639] \\
    7 & GPT-5.4 Mini & Single-Pass & 0.561 & 0.614 & 0.596\, [0.573, 0.620] \\
    8 & Gemini 3.1 Pro Preview & Single-Pass & 0.577 & 0.596 & 0.590\, [0.571, 0.609] \\
    9 & DeepSeek V3.2 & Single-Pass & 0.617 & 0.555 & 0.575\, [0.547, 0.604] \\
    10 & Mistral Small 4 & Single-Pass & 0.616 & 0.482 & 0.527\, [0.510, 0.544] \\
    11 & Llama 4 Scout 17B & Single-Pass & 0.506 & 0.334 & 0.391\, [0.374, 0.409] \\
    \midrule
    1 & ReAct (Sonnet 4.6)$^{*}$ & ReAct & \textbf{0.734} & 0.610 & \textbf{0.652}\, [0.626, 0.687] \\
    2 & ReAct (GPT-5.4) & ReAct & 0.688 & 0.603 & 0.631\, [0.601, 0.663] \\
    3 & ReAct (Gemini 3.1 Pro) & ReAct & 0.723 & 0.511 & 0.582\, [0.556, 0.605] \\
    \midrule
    1 & Parallel Specialists (Sonnet 4.6)$^{*}$ & Parallel Specialists & 0.699 & \textbf{0.621} & 0.647\, [0.626, 0.667] \\
    2 & Parallel Specialists (GPT-5.4) & Parallel Specialists & 0.627 & 0.570 & 0.589\, [0.572, 0.611] \\
    3 & Parallel Specialists (Gemini 3.1 Pro) & Parallel Specialists & 0.607 & 0.505 & 0.539\, [0.518, 0.559] \\
    \midrule
    1 & Triage + Critique (Sonnet 4.6)$^{*}$ & Triage + Critique & 0.675 & 0.599 & 0.624\, [0.606, 0.653] \\
    2 & Triage + Critique (GPT-5.4) & Triage + Critique & 0.592 & 0.540 & 0.557\, [0.537, 0.585] \\
    3 & Triage + Critique (Gemini 3.1 Pro) & Triage + Critique & 0.513 & 0.397 & 0.436\, [0.415, 0.457] \\
    \midrule
    1 & Locator-Extractor (Sonnet 4.6)$^{*}$ & Locator-Extractor & 0.643 & 0.528 & 0.566\, [0.543, 0.592] \\
    2 & Locator-Extractor (GPT-5.4) & Locator-Extractor & 0.523 & 0.492 & 0.502\, [0.481, 0.528] \\
    3 & Locator-Extractor (Gemini 3.1 Pro + GPT-5.4 Mini) & Locator-Extractor & 0.560 & 0.436 & 0.478\, [0.456, 0.502] \\
    4 & Locator-Extractor (Gemini 3.1 Pro) & Locator-Extractor & 0.500 & 0.422 & 0.448\, [0.423, 0.471] \\
    \midrule
    \multicolumn{6}{l}{\textit{Unranked diagnostic: generated the gold pre-fills}} \\
    -- & \textit{Claude Sonnet 4.5}$^{*}$ & \textit{Single-Pass} & \textit{0.903} & \textit{0.840} & \textit{0.861}\, \textit{[0.830, 0.893]} \\
    \bottomrule
    \end{tabular}%
    }
\end{table}

\subsection{Performance Analysis of Evaluated Systems}
\label{sec:headline-ranking}

Table~\ref{tab:test88-headline} reports the composite score for every evaluated system on the test split, broken into Core and RAI strata. Three findings stand out. First, even the strongest system leaves substantial headroom: no configuration exceeds 0.71 composite, and performance on long-form RAI fields trails core fields for most systems, confirming that schema-complete extraction is far from solved. Second, the spread across models is wide (composite scores of 0.39 to 0.71 across all evaluated systems), yet open-weight models such as Qwen~3.6~35B-A3B and GLM-5.1 are competitive with several proprietary models, indicating that capability on this task is not exclusive to frontier systems. Third, agentic architectures underperform single-pass extraction with the same backbone in every case, which we analyze next.

\textbf{Agentic Architectures.} With Claude Sonnet~4.6 fixed as the backbone, all four agentic variants score lower than single-pass: ReAct achieves 0.652 ($\Delta = -0.057$), followed by Parallel Specialists at 0.647 ($\Delta = -0.061$), Triage~+~Critique at 0.624 ($\Delta = -0.085$), and Locator-Extractor at 0.566 ($\Delta = -0.143$). All four differences are statistically significant under both the paired bootstrap and the Wilcoxon signed-rank test after BH-FDR correction ($q < 0.001$). The largest loss comes from Locator-Extractor, the only architecture whose extraction calls see selected passages instead of the full paper; the three architectures that read the full paper lose between 0.057 and 0.085.
The same holds on the other two backbones: single-pass extraction again scores higher than every agentic architecture, by 0.034 to 0.163 on GPT-5.4 and by 0.008 to 0.154 on Gemini~3.1~Pro.

\textbf{Core vs. RAI Fields.} \label{sec:general-vs-rai} We find systems' performance to differ substantially across field types. Most systems score higher on the short, constrained core fields than on RAI fields.
This difference influences the overall performance: RAI fields account for two-thirds of the total score, and therefore largely determine the overall ranking. While Sonnet~4.6 follows the typical pattern (higher core than RAI performance), some models such as Claude Opus~4.7 show slightly better performance on long-form RAI fields than core fields. Two factors likely compound here: (i) Tier~1 rule-based scoring on core fields is exact-match-strict, while the Tier~2 LLM-judge on RAI awards partial credit (0.5) for near-miss answers, so the two columns are not directly commensurable; and (ii) reasoning-tuned backbones (e.g.\ Opus~4.7's extended thinking) gain more on multi-paragraph RAI synthesis than on literal extraction of constrained core values.
Agentic architectures show disproportionately degraded performance on RAI fields. When holding the backbone fixed, all agentic variants show larger performance drops on RAI than on core fields, increasing the gap between the two categories. This suggests that decomposing the task disrupts the cross-section reasoning required for RAI extraction.
The same pattern appears across model sizes. Smaller models perform competitively on core fields but substantially worse on RAI fields, where longer context integration and synthesis are required. The low inter-annotator agreement on long-form RAI fields (Section 3.1) also points to a limitation of the annotation itself, and suggests that more precise field definitions in future schema versions would make both annotation and automated extraction more reliable.

\subsection{Costs \& Error Analysis}
\label{sec:error-analysis}

\textbf{Agentic Decomposition: Failure Mode Analysis}
We trace the gaps between single-pass extraction and the agentic architectures to failure modes introduced by decomposition. Single-pass extraction answers all fields in one call with the whole paper in view, whereas the agentic pipelines spread the work over several calls or steps, and Locator-Extractor also limits each call to selected passages; these pipelines more often return incomplete or incorrect values. This shows up as missed fields: on test-split RAI cells where the gold value is documented in the paper, single-pass extraction returns an empty value for only 2.4\% of cells, whereas the four agentic variants miss between 10.9\% and 14.8\% (Parallel Specialists 10.9\%, Locator-Extractor 12.9\%, ReAct 13.4\%, Triage~+~Critique 14.8\%), a four- to six-fold increase in missed evidence.
Our analysis of 117 sampled cells where agentic systems scored below single-pass Sonnet~4.6 identifies recurring error patterns (Appendix~\ref{app:arch-gap}). Incomplete answers are the most common error for Locator-Extractor (13 of 30 cells). ReAct and Parallel Specialists most often leave documented fields empty, sometimes despite reporting the relevant information elsewhere. In 5 of the 13 empty Parallel Specialists cells, another specialist places the information under a different field. In 8 of the 15 empty ReAct cells, the agent reports the information under another field or mentions it in its explanation for leaving the target field empty. Triage~+~Critique leaves the most documented fields empty (14.8\%). This follows from its deliberately cautious design: the extraction prompt asks the model to fill a field only when the paper discusses it explicitly, and the verification step keeps long-form answers only when their supporting quote is found in the paper. These safeguards are meant to prevent unsupported answers, at the cost of sometimes leaving documented fields empty.

\begin{figure}[!t]
  \centering
  \includegraphics[width=\textwidth]{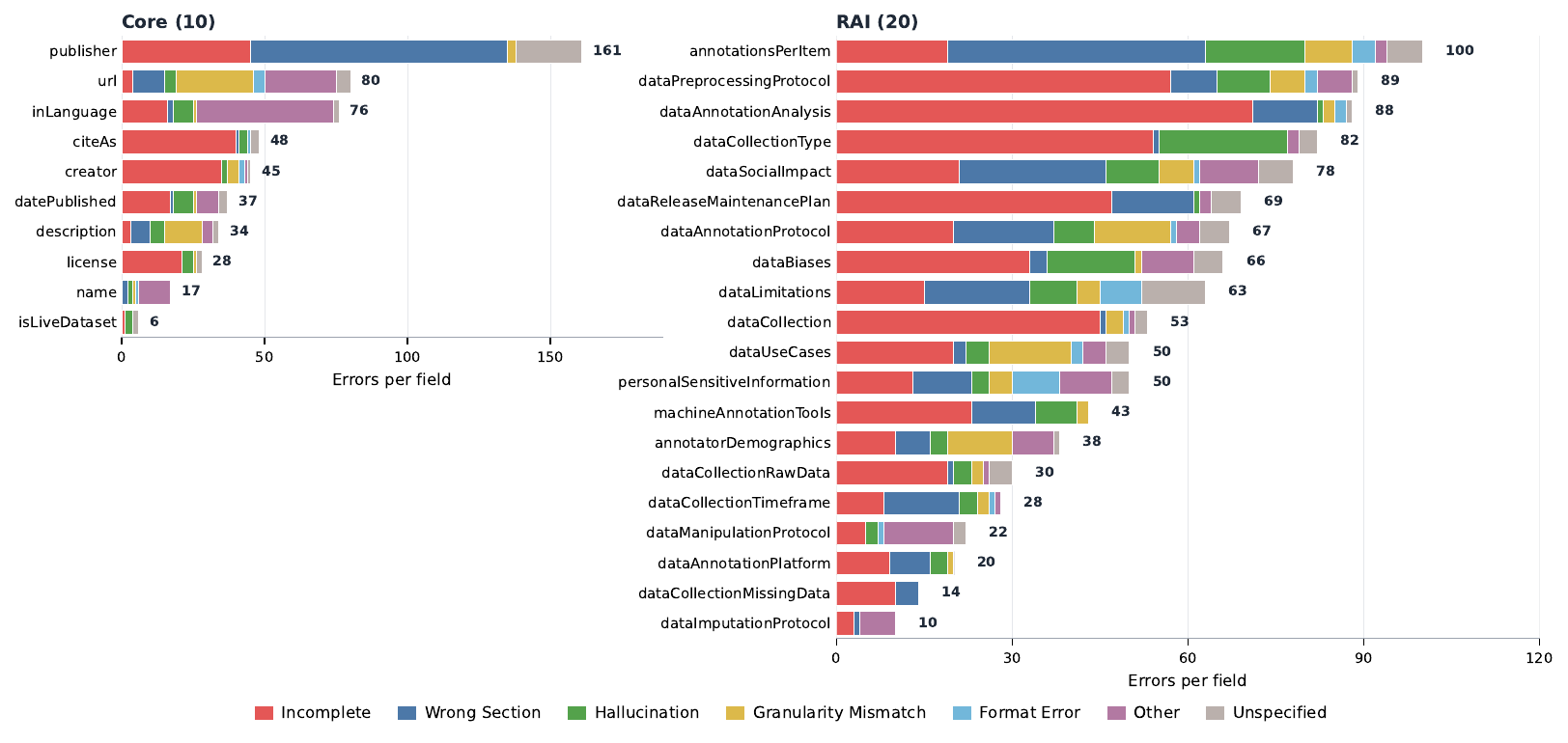}
  \caption{Failure-mode taxonomy across 1{,}592 error ratings. Cell counts per
    (field, failure-mode) bin; rows sorted by total descending within
    each panel. \emph{Incomplete} extraction dominates (43\%) for long-form RAI fields. The single counter-pattern
    is \texttt{sc:publisher}, where \emph{Wrong Section} dominates:
    annotators disagree about whether the dataset host, venue, or author affiliation counts
    as the publisher.}
  \label{fig:failure-modes}
\end{figure}

\textbf{Aggregate Patterns.} Among the 16.6\% of pre-fill ratings that annotators flagged (Section~\ref{sec:annotation}), incomplete extraction (43.0\%) and wrong-section assignment (19.5\%) dominate, while hallucinations account for only 9.7\% of errors. Models often extract valid information from the paper but assign it to the wrong field, indicating that errors are often due to misclassification rather than fabrication (see Fig.~\ref{fig:failure-modes}).

\textbf{Field-level Trends.} Errors are unevenly distributed across fields and are particularly frequent for a small subset of RAI attributes that require multi-step reasoning or aggregation across sections such as \texttt{rai:dataPreprocessingProtocol} and \texttt{rai:dataAnnotationAnalysis} (Fig.~\ref{fig:failure-modes}). In contrast, structured core fields are generally extracted reliably. For single-pass extraction, a paper's difficulty depends on how sparsely it is documented rather than on its length: scores fall steadily from the best-documented to the most sparsely documented papers, while page count has no measurable effect (Appendix~\ref{app:difficulty}).

\textbf{Implications.} These patterns suggest that improving extraction quality does not simply require stronger models, but better handling of structure and uncertainty. Concretely, encouraging systems to leave a field empty when the evidence is unclear, validating outputs against field definitions, and refining ambiguous schema fields can reduce errors. Addressing schema-level ambiguities (e.g., the definition of \texttt{publisher}) is necessary for consistent evaluation and further progress.

\textbf{Costs.} Single-pass extraction is the most cost-efficient variant across all backbone-models. At standard list prices and under the token-accounting assumptions in Appendix~\ref{app:cost}, agentic configurations cost 1.2$\times$ to 7.8$\times$ as much per paper as single-pass extraction in matched-backbone comparisons, without improving composite scores. Combined with the accuracy results in Sections~\ref{sec:headline-ranking} and~\ref{sec:error-analysis}, this makes agentic approaches suboptimal across all budget tiers we evaluate. We report per-system costs in Appendix~\ref{app:cost}.

\section{Conclusion}
\label{sec:conclusion}

We presented \cm, a benchmark for automated extraction of Croissant metadata from ML dataset papers, and evaluated 24 extraction systems on it. The \cm benchmark introduces 602 dataset papers annotated with the full 30-field Croissant schema, enabling the first end-to-end evaluation of metadata extraction on realistic dataset documentation. Across all evaluated systems, single-pass extraction consistently outperforms the agentic architectures we tested. This gap is most pronounced for long-form RAI fields, where accurate extraction requires integrating information across multiple sections of a paper. Beyond establishing a new benchmark, our results highlight both the potential and current limitations of LLM-based metadata extraction, and provide a foundation for future work on scalable, reliable dataset documentation.

\section*{Limitations}
\label{sec:limitations}

First, our benchmark measures how well systems extract what papers document, and this entangles two sources of difficulty: extraction capability and the underlying incompleteness of dataset documentation. In the gold split, 27.7\% of cells have no documented value, so part of every composite score reflects whether a system correctly leaves undocumented fields empty, not only how well it extracts; low scores on sparsely documented fields should not be read as extraction failures alone. Second, Tier-2 scores rely on an LLM judge. Its agreement with human ratings is imperfect, especially on difficult fields (Section~\ref{sec:experimental-setup}), so judge-based scores do not replace a full human evaluation. Third, the gold labels began as LLM pre-fills. Re-annotating 10 papers from a different model's pre-fills left the ranking stable but favoured the seed model and, slightly, its family (Appendix~\ref{app:seed-robustness}), so some anchoring to the pre-fill model cannot be ruled out. Fourth, several RAI fields have very few documented cells in the test split, as few as two for \texttt{rai:dataImputationProtocol}, so results for these fields should not be over-read, and although scores show no relation to dataset popularity or publication date (Appendix~\ref{app:difficulty}), pretraining contamination cannot be ruled out. Finally, the benchmark covers English-language ML dataset papers and the 30-field Croissant 1.1 schema; extension to other languages, domains, and future schema versions is left to future work. These extensions and the preceding limitations are discussed further in Appendix~\ref{app:limitations}, which also examines caveats concerning PDF parsing, model non-determinism, and architecture coverage.

\begin{ack}
We thank our annotators for their careful work in building the gold benchmark: Aman Krishnatrey, Amelia Jim\'enez-S\'anchez, Aneesha Das, Anwai Archit, Cory Levinson, Diwakar Gupta, Kyoungsook Kim, Luis Oala, Manoj Devender, My Chiffon Nguyen, Shriyash Gulhane, Taanish Bhardwaj, Varuni H K, and Vitor Basto Fernandes. We acknowledge the MLCommons Croissant Working Group for developing the metadata schema, and the ETH Zurich AI Center and LRE group for institutional support. Compute resources for self-hosted open-weight runs were provided by the ETH Zurich Euler cluster. Mubashara Akhtar was primarily supported by the ETH AI Center through an ETH AI Center postdoctoral fellowship. Joan Giner-Miguelez acknowledges their AI4S fellowship from the Ministerio para la Transformación Digital y de la Función Pública, for talent attraction (C005/24-ED CV1), funded by NextGenerationEU through PRTR and has received funding from the European Union under the AI4SOCIAL+ Project, grant agreement 101292886.
\end{ack}

\bibliographystyle{plainnat}
\bibliography{references}

\begin{thebibliography}{59}
\providecommand{\natexlab}[1]{#1}
\providecommand{\url}[1]{\texttt{#1}}
\expandafter\ifx\csname urlstyle\endcsname\relax
  \providecommand{\doi}[1]{doi: #1}\else
  \providecommand{\doi}{doi: \begingroup \urlstyle{rm}\Url}\fi

\bibitem[Akhtar et~al.(2024)Akhtar, Benjelloun, Conforti, Foschini, Gijsbers, Giner-Miguelez, Goswami, Jain, Karamousadakis, Krishna, et~al.]{akhtar2024croissant}
Mubashara Akhtar, Omar Benjelloun, Costanza Conforti, Luca Foschini, Pieter Gijsbers, Joan Giner-Miguelez, Sujata Goswami, Nitisha Jain, Michalis Karamousadakis, Satyapriya Krishna, et~al.
\newblock Croissant: A metadata format for {ML}-ready datasets.
\newblock In \emph{Advances in Neural Information Processing Systems}, volume~37, pages 82133--82148, 2024.

\bibitem[Alyafeai et~al.(2025{\natexlab{a}})Alyafeai, Al-Shaibani, and Ghanem]{alyafeai2025mextract}
Zaid Alyafeai, Maged~S Al-Shaibani, and Bernard Ghanem.
\newblock Mextract: Light-weight metadata extraction from scientific papers.
\newblock \emph{arXiv preprint arXiv:2510.06889}, 2025{\natexlab{a}}.

\bibitem[Alyafeai et~al.(2025{\natexlab{b}})Alyafeai, AlShaibani, and Ghanem]{alyafeai2025mole}
Zaid Alyafeai, Maged~Saeed AlShaibani, and Bernard Ghanem.
\newblock {MOLE}: Metadata extraction and validation in scientific papers using {LLMs}.
\newblock In \emph{Findings of the Association for Computational Linguistics: EMNLP 2025}, pages 12236--12264, 2025{\natexlab{b}}.

\bibitem[Bender and Friedman(2018)]{bender2018data}
Emily~M Bender and Batya Friedman.
\newblock Data statements for natural language processing: Toward mitigating system bias and enabling better science.
\newblock \emph{Transactions of the Association for Computational Linguistics}, 6:\penalty0 587--604, 2018.

\bibitem[Benjamini and Hochberg(1995)]{benjamini1995controlling}
Yoav Benjamini and Yosef Hochberg.
\newblock Controlling the false discovery rate: A practical and powerful approach to multiple testing.
\newblock \emph{Journal of the Royal Statistical Society, Series B (Methodological)}, 57\penalty0 (1):\penalty0 289--300, 1995.

\bibitem[Bhardwaj et~al.(2024)Bhardwaj, Gujral, Wu, Zogheib, Maharaj, and Becker]{bhardwaj2024machine}
Eshta Bhardwaj, Harshit Gujral, Siyi Wu, Ciara Zogheib, Tegan Maharaj, and Christoph Becker.
\newblock Machine learning data practices through a data curation lens: An evaluation framework.
\newblock In \emph{Proceedings of the 2024 ACM Conference on Fairness, Accountability, and Transparency}, pages 1055--1067, 2024.

\bibitem[Dagdelen et~al.(2024)Dagdelen, Dunn, Lee, Walker, Rosen, Ceder, Persson, and Jain]{dagdelen2024structured}
John Dagdelen, Alexander Dunn, Sanghoon Lee, Nicholas Walker, Andrew~S Rosen, Gerbrand Ceder, Kristin~A Persson, and Anubhav Jain.
\newblock Structured information extraction from scientific text with large language models.
\newblock \emph{Nature Communications}, 15:\penalty0 1418, 2024.

\bibitem[D{\'\i}az et~al.(2022)D{\'\i}az, Kivlichan, Rosen, Baker, Amironesei, Prabhakaran, and Denton]{diaz2022crowdworksheets}
Mark D{\'\i}az, Ian Kivlichan, Rachel Rosen, Dylan Baker, Razvan Amironesei, Vinodkumar Prabhakaran, and Remi Denton.
\newblock Crowdworksheets: Accounting for individual and collective identities underlying crowdsourced dataset annotation.
\newblock In \emph{Proceedings of the 2022 ACM Conference on Fairness, Accountability, and Transparency}, pages 2342--2351, 2022.

\bibitem[Dodge et~al.(2021)Dodge, Sap, Marasovi{\'c}, Agnew, Ilharco, Groeneveld, Mitchell, and Gardner]{dodge2021documenting}
Jesse Dodge, Maarten Sap, Ana Marasovi{\'c}, William Agnew, Gabriel Ilharco, Dirk Groeneveld, Margaret Mitchell, and Matt Gardner.
\newblock Documenting large webtext corpora: A case study on the colossal clean crawled corpus.
\newblock In \emph{Proceedings of the 2021 conference on empirical methods in natural language processing}, pages 1286--1305, 2021.

\bibitem[Efron(1979)]{efron1979bootstrap}
Bradley Efron.
\newblock Bootstrap methods: Another look at the jackknife.
\newblock \emph{The Annals of Statistics}, 7\penalty0 (1):\penalty0 1--26, 1979.

\bibitem[{European Parliament and Council of the European Union}(2024)]{euaiact2024art11}
{European Parliament and Council of the European Union}.
\newblock Regulation (eu) 2024/1689 laying down harmonised rules on artificial intelligence (artificial intelligence act), article 11: Technical documentation.
\newblock \url{https://eur-lex.europa.eu/legal-content/EN/TXT/?uri=CELEX:32024R1689}, 2024.
\newblock Official Journal of the European Union, L 2024/1689, 12 July 2024; entered into force 1 August 2024.

\bibitem[Ferguson et~al.(2026)Ferguson, Pennington, Beghian, Mohan, Kiela, Agrawal, and Nguyen]{ferguson2026extractbench}
Nick Ferguson, Josh Pennington, Narek Beghian, Aravind Mohan, Douwe Kiela, Sheshansh Agrawal, and Thien~Hang Nguyen.
\newblock Extractbench: A benchmark and evaluation methodology for complex structured extraction.
\newblock \emph{arXiv preprint arXiv:2602.12247}, 2026.

\bibitem[Gebru et~al.(2021)Gebru, Morgenstern, Vecchione, Vaughan, Wallach, Daum{\'e}~III, and Crawford]{gebru2021datasheets}
Timnit Gebru, Jamie Morgenstern, Briana Vecchione, Jennifer~Wortman Vaughan, Hanna Wallach, Hal Daum{\'e}~III, and Kate Crawford.
\newblock Datasheets for datasets.
\newblock \emph{Communications of the ACM}, 64\penalty0 (12):\penalty0 86--92, 2021.

\bibitem[Ghosh et~al.(2026)Ghosh, Kaffee, Jernite, and Solaiman]{hf_state_of_os_spring26}
Avijit Ghosh, Lucie-Aimée Kaffee, Yacine Jernite, and Irene Solaiman.
\newblock State of open source on hugging face: Spring 2026.
\newblock \url{https://huggingface.co/blog/huggingface/state-of-os-hf-spring-2026}, 2026.
\newblock Hugging Face Blog, Accessed: 2026-04-11.

\bibitem[Giner-Miguelez et~al.(2023)Giner-Miguelez, G{\'o}mez, and Cabot]{giner2023datadoc}
Joan Giner-Miguelez, Abel G{\'o}mez, and Jordi Cabot.
\newblock Datadoc analyzer: A tool for analyzing the documentation of scientific datasets.
\newblock In \emph{Proceedings of the 32nd ACM International Conference on Information and Knowledge Management}, pages 5046--5050, 2023.

\bibitem[Holland et~al.(2020)Holland, Hosny, Newman, Joseph, and Chmielinski]{holland2018dataset}
Sarah Holland, Ahmed Hosny, Sarah Newman, Joshua Joseph, and Kasia Chmielinski.
\newblock The dataset nutrition label: A framework to drive higher data quality standards.
\newblock In Dara Hallinan, Ronald Leenes, Serge Gutwirth, and Paul De~Hert, editors, \emph{Data Protection and Privacy, Volume 12: Data Protection and Democracy}, chapter~1. Hart Publishing, 2020.
\newblock ISBN 978-1-50993-274-0.
\newblock \doi{10.5040/9781509932771.ch-001}.

\bibitem[{Hugging Face Community}(2026)]{hfhubstats2026}
{Hugging Face Community}.
\newblock Hugging face hub statistics.
\newblock \url{https://huggingface.co/spaces/cfahlgren1/hub-stats}, 2026.
\newblock Accessed: 2026-04-11.

\bibitem[Jain et~al.(2024)Jain, Akhtar, Giner-Miguelez, Shinde, Vanschoren, Vogler, Goswami, Rao, Santos, Oala, et~al.]{jain2024standardized}
Nitisha Jain, Mubashara Akhtar, Joan Giner-Miguelez, Rajat Shinde, Joaquin Vanschoren, Steffen Vogler, Sujata Goswami, Yuhan Rao, Tim Santos, Luis Oala, et~al.
\newblock A standardized machine-readable dataset documentation format for responsible {AI}.
\newblock \emph{arXiv preprint arXiv:2407.16883}, 2024.

\bibitem[Jain et~al.(2020)Jain, van Zuylen, Hajishirzi, and Beltagy]{jain2020scirex}
Sarthak Jain, Madeleine van Zuylen, Hannaneh Hajishirzi, and Iz~Beltagy.
\newblock {SciREX}: A challenge dataset for document-level information extraction.
\newblock In \emph{Proceedings of ACL}, 2020.

\bibitem[Jia et~al.(2025)Jia, Oala, Xiong, Ge, Wang, Kang, and Song]{jia2025sustainable}
Ruoxi Jia, Luis Oala, Wenjie Xiong, Suqin Ge, Jiachen~T Wang, Feiyang Kang, and Dawn Song.
\newblock A sustainable ai economy needs data deals that work for generators.
\newblock In \emph{The Thirty-Ninth Annual Conference on Neural Information Processing Systems Position Paper Track}, 2025.

\bibitem[Kim et~al.(2025)Kim, Sohn, Jo, Choi, Bae, Lee, Park, and Lee]{kim2025not}
Jaekyeom Kim, Sungryull Sohn, Gerrard~Jeongwon Jo, Jihoon Choi, Kyunghoon Bae, Hwayoung Lee, Yongmin Park, and Honglak Lee.
\newblock Do not trust licenses you see: Dataset compliance requires massive-scale ai-powered lifecycle tracing.
\newblock \emph{arXiv preprint arXiv:2503.02784}, 2025.

\bibitem[Lhoest et~al.(2021)Lhoest, Del~Moral, Jernite, Thakur, Von~Platen, Patil, Chaumond, Drame, Plu, Tunstall, et~al.]{lhoest2021datasets}
Quentin Lhoest, Albert~Villanova Del~Moral, Yacine Jernite, Abhishek Thakur, Patrick Von~Platen, Suraj Patil, Julien Chaumond, Mariama Drame, Julien Plu, Lewis Tunstall, et~al.
\newblock Datasets: A community library for natural language processing.
\newblock In \emph{Proceedings of the 2021 conference on empirical methods in natural language processing: system demonstrations}, pages 175--184, 2021.

\bibitem[Liang et~al.(2025)Liang, Zhang, Wu, Tang, and Liu]{liang2025schema}
Sheng Liang, Yongyue Zhang, Yaxiong Wu, Ruiming Tang, and Yong Liu.
\newblock Schema as parameterized tools for universal information extraction.
\newblock \emph{arXiv preprint arXiv:2506.01276}, 2025.

\bibitem[Liu et~al.(2024{\natexlab{a}})Liu, Li, Jin, and Diab]{liu2024cardgen}
Jiarui Liu, Wenkai Li, Zhijing Jin, and Mona Diab.
\newblock Automatic generation of model and data cards: A step towards responsible {AI}.
\newblock In \emph{Proceedings of the 2024 Conference of the North American Chapter of the Association for Computational Linguistics (NAACL)}, 2024{\natexlab{a}}.

\bibitem[Liu et~al.(2024{\natexlab{b}})Liu, Wei, Liu, Si, Zhang, Rao, Zheng, Peng, Yang, Zhou, et~al.]{liu2024best}
Ruibo Liu, Jerry Wei, Fangyu Liu, Chenglei Si, Yanzhe Zhang, Jinmeng Rao, Steven Zheng, Daiyi Peng, Diyi Yang, Denny Zhou, et~al.
\newblock Best practices and lessons learned on synthetic data.
\newblock In \emph{First Conference on Language Modeling}, 2024{\natexlab{b}}.

\bibitem[Liu et~al.(2023)Liu, Iter, Xu, Wang, Xu, and Zhu]{liu2023geval}
Yang Liu, Dan Iter, Yichong Xu, Shuohang Wang, Ruochen Xu, and Chenguang Zhu.
\newblock {G-Eval}: {NLG} evaluation using {GPT-4} with better human alignment.
\newblock In \emph{Proceedings of EMNLP}, 2023.

\bibitem[Longpre et~al.(2024)Longpre, Mahari, Chen, Obeng-Marnu, Sileo, Brannon, Muennighoff, Khazam, Kabbara, Perisetla, et~al.]{longpre2024large}
Shayne Longpre, Robert Mahari, Anthony Chen, Naana Obeng-Marnu, Damien Sileo, William Brannon, Niklas Muennighoff, Nathan Khazam, Jad Kabbara, Kartik Perisetla, et~al.
\newblock A large-scale audit of dataset licensing and attribution in ai.
\newblock \emph{Nature Machine Intelligence}, pages 975--987, 2024.

\bibitem[Luan et~al.(2018)Luan, He, Ostendorf, and Hajishirzi]{luan2018multitask}
Yi~Luan, Luheng He, Mari Ostendorf, and Hannaneh Hajishirzi.
\newblock Multi-task identification of entities, relations, and coreference for scientific knowledge graph construction.
\newblock In \emph{Proceedings of EMNLP}, 2018.

\bibitem[Luccioni et~al.(2022)Luccioni, Corry, Sridharan, Ananny, Schultz, and Crawford]{luccioni2022framework}
Alexandra~Sasha Luccioni, Frances Corry, Hamsini Sridharan, Mike Ananny, Jason Schultz, and Kate Crawford.
\newblock A framework for deprecating datasets: Standardizing documentation, identification, and communication.
\newblock In \emph{Proceedings of the 2022 acm conference on fairness, accountability, and transparency}, pages 199--212, 2022.

\bibitem[Marini et~al.(2025)Marini, Santos, Contaxis, and Freire]{marini2025data}
Pietro Marini, A{\'e}cio Santos, Nicole Contaxis, and Juliana Freire.
\newblock Data gatherer: Llm-powered dataset reference extraction from scientific literature.
\newblock In \emph{Proceedings of the Fifth Workshop on Scholarly Document Processing (SDP 2025)}, pages 114--123, 2025.

\bibitem[McNemar(1947)]{mcnemar1947note}
Quinn McNemar.
\newblock Note on the sampling error of the difference between correlated proportions or percentages.
\newblock \emph{Psychometrika}, 12\penalty0 (2):\penalty0 153--157, 1947.

\bibitem[Mitchell et~al.(2019)Mitchell, Wu, Zaldivar, Barnes, Vasserman, Hutchinson, Spitzer, Raji, and Gebru]{mitchell2019model}
Margaret Mitchell, Simone Wu, Andrew Zaldivar, Parker Barnes, Lucy Vasserman, Ben Hutchinson, Elena Spitzer, Inioluwa~Deborah Raji, and Timnit Gebru.
\newblock Model cards for model reporting.
\newblock In \emph{Proceedings of the Conference on Fairness, Accountability, and Transparency}, pages 220--229, 2019.

\bibitem[{MLCommons}(2025)]{mlcommons2025croissant}
{MLCommons}.
\newblock Croissant: A metadata format for machine learning datasets.
\newblock \url{https://mlcommons.org/working-groups/data/croissant/}, 2025.
\newblock MLCommons Working Group, accessed: 2026-04-11.

\bibitem[{NeurIPS Communications Chairs}(2025)]{neurips2025db}
{NeurIPS Communications Chairs}.
\newblock Neurips datasets \& benchmarks track: From art to science in ai evaluations.
\newblock \url{https://blog.neurips.cc/2025/12/05/neurips-datasets-benchmarks-track-from-art-to-science-in-ai-evaluations/}, 2025.
\newblock NeurIPS Blog, accessed: 2026-04-11.

\bibitem[{NeurIPS Organizing Committee}(2026)]{neurips2026edfaq}
{NeurIPS Organizing Committee}.
\newblock Neurips 2026 evaluations \& datasets faq.
\newblock \url{https://neurips.cc/Conferences/2026/EvaluationsDatasetsFAQ}, 2026.
\newblock accessed: 2026-04-11.

\bibitem[{NuMind}(2024)]{numind2024nuextract}
{NuMind}.
\newblock {NuExtract}: A foundation model for structured extraction, 2024.
\newblock URL \url{https://numind.ai/blog/nuextract-a-foundation-model-for-structured-extraction}.

\bibitem[Oren et~al.(2024)Oren, Meister, Chatterji, Ladhak, and Hashimoto]{oren2023proving}
Yonatan Oren, Nicole Meister, Niladri Chatterji, Faisal Ladhak, and Tatsunori~B. Hashimoto.
\newblock Proving test set contamination in black box language models.
\newblock In \emph{International Conference on Learning Representations (ICLR)}, 2024.

\bibitem[Peng et~al.(2021)Peng, Mathur, and Narayanan]{peng2021mitigating}
Kenny Peng, Arunesh Mathur, and Arvind Narayanan.
\newblock Mitigating dataset harms requires stewardship: Lessons from 1000 papers.
\newblock In \emph{Thirty-fifth Conference on Neural Information Processing Systems Datasets and Benchmarks Track (Round 2)}, 2021.

\bibitem[Perot et~al.(2024)Perot, Kang, Luisier, Su, Sun, Boppana, Wang, Wang, Mu, Zhang, et~al.]{perot2024lmdx}
Vincent Perot, Kai Kang, Florian Luisier, Guolong Su, Xiaoyu Sun, Ramya~Sree Boppana, Zilong Wang, Zifeng Wang, Jiaqi Mu, Hao Zhang, et~al.
\newblock Lmdx: Language model-based document information extraction and localization.
\newblock In \emph{Findings of the Association for Computational Linguistics: ACL 2024}, pages 15140--15168, 2024.

\bibitem[Polak and Morgan(2024)]{polak2024extracting}
Maciej~P Polak and Dane Morgan.
\newblock Extracting accurate materials data from research papers with conversational language models and prompt engineering.
\newblock \emph{Nature Communications}, 15:\penalty0 1569, 2024.

\bibitem[Pushkarna et~al.(2022)Pushkarna, Zaldivar, and Kjartansson]{pushkarna2022data}
Mahima Pushkarna, Andrew Zaldivar, and Oddur Kjartansson.
\newblock Data cards: Purposeful and transparent dataset documentation for responsible {AI}.
\newblock In \emph{Proceedings of the 2022 ACM Conference on Fairness, Accountability, and Transparency}, pages 1776--1826, 2022.

\bibitem[Reynolds-Cu{\'e}llar et~al.(2026)Reynolds-Cu{\'e}llar, Wong-Villacres, Alvarado~Garcia, and Precel]{reynolds2026reflection}
Pedro Reynolds-Cu{\'e}llar, Marisol Wong-Villacres, Adriana Alvarado~Garcia, and Heila Precel.
\newblock From reflection to repair: A scoping review of dataset documentation tools.
\newblock In \emph{Proceedings of the 2026 CHI Conference on Human Factors in Computing Systems}, pages 1--25, 2026.

\bibitem[Sainz et~al.(2024)Sainz, Garc{\'\i}a-Ferrero, Agerri, de~Lacalle, Rigau, and Agirre]{sainz2024gollie}
Oscar Sainz, Iker Garc{\'\i}a-Ferrero, Rodrigo Agerri, Oier~Lopez de~Lacalle, German Rigau, and Eneko Agirre.
\newblock Go{LLIE}: Annotation guidelines improve zero-shot information-extraction.
\newblock In \emph{The Twelfth International Conference on Learning Representations}, volume 2024, pages 47083--47107, 2024.

\bibitem[Sambasivan et~al.(2021)Sambasivan, Kapania, Highfill, Akrong, Paritosh, and Aroyo]{sambasivan2021everyone}
Nithya Sambasivan, Shivani Kapania, Hannah Highfill, Diana Akrong, Praveen Paritosh, and Lora~M Aroyo.
\newblock ``{E}veryone wants to do the model work, not the data work'': Data cascades in high-stakes {AI}.
\newblock In \emph{proceedings of the 2021 CHI Conference on Human Factors in Computing Systems}, pages 1--15, 2021.

\bibitem[Seidenberger and Maiti(2025)]{seidenberger2025big}
Scott Seidenberger and Anindya Maiti.
\newblock From big data to valued data: A dataset value taxonomy for ai-native empirical research.
\newblock In \emph{Proceedings of the AAAI/ACM Conference on AI, Ethics, and Society}, pages 2294--2305, 2025.

\bibitem[Shrimal et~al.(2025)Shrimal, Jain, Chowdhury, and Yenigalla]{shrimal2025parse}
Anubhav Shrimal, Aryan Jain, Soumyajit Chowdhury, and Promod Yenigalla.
\newblock Parse: Llm driven schema optimization for reliable entity extraction.
\newblock In \emph{Proceedings of the 2025 Conference on Empirical Methods in Natural Language Processing: Industry Track}, pages 2749--2763, 2025.

\bibitem[Tenckhoff et~al.(2026)Tenckhoff, Koddenbrock, and Rodner]{tenckhoff2026llmstructbench}
S{\"o}nke Tenckhoff, Mario Koddenbrock, and Erik Rodner.
\newblock Llmstructbench: Benchmarking large language model structured data extraction.
\newblock \emph{arXiv preprint arXiv:2602.14743}, 2026.

\bibitem[Vanschoren et~al.(2014)Vanschoren, van Rijn, Bischl, and Torgo]{vanschoren2014openml}
Joaquin Vanschoren, Jan~N van Rijn, Bernd Bischl, and Lu{\'\i}s Torgo.
\newblock {OpenML}: Networked science in machine learning.
\newblock \emph{ACM SIGKDD Explorations}, 15\penalty0 (2):\penalty0 49--60, 2014.

\bibitem[Wang et~al.(2023)Wang, Wei, Schuurmans, Le, Chi, Narang, Chowdhery, and Zhou]{wang2023selfconsistency}
Xuezhi Wang, Jason Wei, Dale Schuurmans, Quoc Le, Ed~Chi, Sharan Narang, Aakanksha Chowdhery, and Denny Zhou.
\newblock Self-consistency improves chain of thought reasoning in language models.
\newblock In \emph{Proceedings of ICLR}, 2023.

\bibitem[Wilcoxon(1945)]{wilcoxon1945individual}
Frank Wilcoxon.
\newblock Individual comparisons by ranking methods.
\newblock \emph{Biometrics Bulletin}, 1\penalty0 (6):\penalty0 80--83, 1945.

\bibitem[Willard and Louf(2023)]{willard2023efficient}
Brandon~T Willard and R{\'e}mi Louf.
\newblock Efficient guided generation for large language models.
\newblock \emph{arXiv preprint arXiv:2307.09702}, 2023.

\bibitem[Winata et~al.(2025)Winata, Anugraha, Liu, Aji, Hung, Parashar, Irawan, Zhang, Yong, Cruz, et~al.]{winata2025datasheets}
Genta~Indra Winata, David Anugraha, Emmy Liu, Alham~Fikri Aji, Shou-Yi Hung, Aditya Parashar, Patrick~Amadeus Irawan, Ruochen Zhang, Zheng-Xin Yong, Jan Christian~Blaise Cruz, et~al.
\newblock Datasheets aren't enough: Datarubrics for automated quality metrics and accountability.
\newblock \emph{arXiv preprint arXiv:2506.01789}, 2025.

\bibitem[Wu et~al.(2024)Wu, Ajmani, Longpre, and Li]{wu2024systematic}
Yiwei Wu, Leah Ajmani, Shayne Longpre, and Hanlin Li.
\newblock A systematic review of neurips dataset management practices.
\newblock \emph{Advances in Neural Information Processing Systems}, 37:\penalty0 32813--32827, 2024.

\bibitem[Yang et~al.(2026)Yang, Jiang, He, Siu, Zhang, Liao, Li, Zeng, Jia, Wang, Schneider, Ruan, Ma, Lyu, Wang, Lu, Do, Jiang, Nie, and Chen]{yang2026structeval}
Jialin Yang, Dongfu Jiang, Tony He, Sherman Siu, Yuxuan Zhang, Disen Liao, Zhuofeng Li, Huaye Zeng, Yiming Jia, Haozhe Wang, Benjamin Schneider, Chi Ruan, Wentao Ma, Zhiheng Lyu, Yifei Wang, Yi~Lu, Quy~Duc Do, Ziyan Jiang, Ping Nie, and Wenhu Chen.
\newblock Structeval: Benchmarking {LLM}s' capabilities to generate structural outputs.
\newblock \emph{Transactions on Machine Learning Research}, 2026.
\newblock ISSN 2835-8856.

\bibitem[Yang et~al.(2024)Yang, Liang, and Zou]{yang2024navigating}
Xinyu Yang, Victor~Weixin Liang, and James~Y Zou.
\newblock Navigating dataset documentations in {AI}: A large-scale analysis of dataset cards on huggingface.
\newblock In \emph{International Conference on Learning Representations}, pages 33011--33028, 2024.

\bibitem[Yao et~al.(2023)Yao, Zhao, Yu, Du, Shafran, Narasimhan, and Cao]{yao2023react}
Shunyu Yao, Jeffrey Zhao, Dian Yu, Nan Du, Izhak Shafran, Karthik Narasimhan, and Yuan Cao.
\newblock {ReAct}: Synergizing reasoning and acting in language models.
\newblock In \emph{International Conference on Learning Representations (ICLR)}, 2023.

\bibitem[Zha et~al.(2025)Zha, Bhat, Lai, Yang, Jiang, Zhong, and Hu]{zha2025data}
Daochen Zha, Zaid~Pervaiz Bhat, Kwei-Herng Lai, Fan Yang, Zhimeng Jiang, Shaochen Zhong, and Xia Hu.
\newblock Data-centric artificial intelligence: A survey.
\newblock \emph{ACM Computing Surveys}, 57\penalty0 (5):\penalty0 1--42, 2025.

\bibitem[Zhao et~al.(2024)Zhao, Scheuerman, Chitre, Andrews, Panagiotidou, Walker, Pine, and Xiang]{zhao2024taxonomy}
Dora Zhao, Morgan~K Scheuerman, Pooja Chitre, Jerone~T Andrews, Georgia Panagiotidou, Shawn Walker, Kathleen~H Pine, and Alice Xiang.
\newblock A taxonomy of challenges to curating fair datasets.
\newblock \emph{Advances in Neural Information Processing Systems}, 37:\penalty0 97826--97858, 2024.

\bibitem[Zheng et~al.(2023)Zheng, Chiang, Sheng, Zhuang, Wu, Zhuang, Lin, Li, Li, Xing, et~al.]{zheng2023judging}
Lianmin Zheng, Wei-Lin Chiang, Ying Sheng, Siyuan Zhuang, Zhanghao Wu, Yonghao Zhuang, Zi~Lin, Zhuohan Li, Dacheng Li, Eric~P Xing, et~al.
\newblock Judging {LLM}-as-a-judge with {MT-Bench} and chatbot arena.
\newblock In \emph{Advances in Neural Information Processing Systems}, volume~36, 2023.

\end{thebibliography}

\appendix

\FloatBarrier
\newpage

\section{Dataset Construction and Annotation Details}

\subsection{Croissant Field Definitions}
\label{app:fields}

\begin{figure}[!ht]
  \centering
  \includegraphics[width=0.9\linewidth]{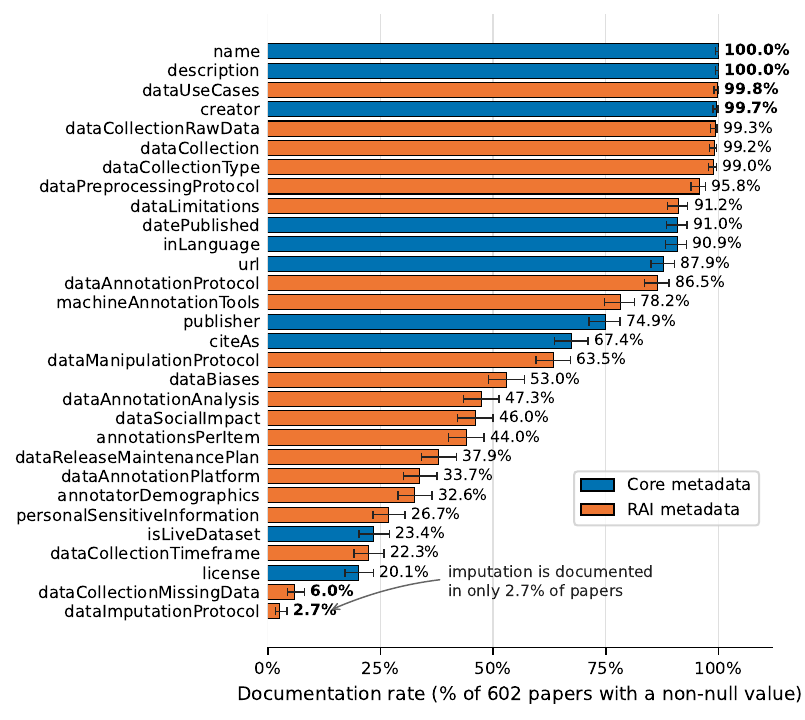}
  \caption{Croissant field documentation rates across the 602-paper \cm benchmark. Each bar shows the percentage of papers with a non-null value for the field. Core fields are consistently reported (often higher than 99\%), while RAI fields show substantial variability, ranging from widely documented (e.g., use cases, limitations) to rarely reported (e.g., imputation, missing data). This highlights a large gap between core metadata coverage and RAI documentation in current dataset papers.}
  \label{fig:field_fill_rates}
\end{figure}

Table~\ref{tab:field-overview} lists all 30 Croissant 1.1 metadata
fields with their schema category (Core vs RAI), the metric used
to score them in this paper (constrained, short-text, or LLM judge),
a one-line definition, and the percentage of papers in which the
field has a non-null gold value (Gold, $N{=}102$) or silver value
(Silver, $N{=}500$). The metric column maps to the two-tier scoring
pipeline of Section~\ref{sec:experimental-setup}: the seven
constrained and three short-text fields are scored by Tier~1
rule-based metrics (exact-match for enums, token-F1 for short text);
the twenty long-form RAI fields are scored by the Tier~2 GLM-5 LLM
judge. Per-field documentation rates over all 602 papers are visualised in
Figure~\ref{fig:field_fill_rates}.

\begin{table*}[t]
    \centering
    \caption{Overview of the 30 metadata fields in the Croissant 1.1 schema (10 core + 20 RAI). Coverage shows the percentage of datasets where each field has a non-null value in the human gold (N=102) and in the silver extraction (N=500).}
    \label{tab:field-overview}
    \footnotesize
    \begin{tabular}{p{4.2cm}ll>{\raggedright\arraybackslash}p{3.6cm}rr}
    \toprule
    Field & Type & Metric & Definition & Gold & Silver \\
    \midrule
    \texttt{name} & Core & Constrained & Dataset name & 100\% & 100\% \\
    \texttt{description} & Core & Short-text & Brief dataset description & 100\% & 100\% \\
    \texttt{url} & Core & Constrained & Access URL & 96\% & 86\% \\
    \texttt{license} & Core & Constrained & Distribution license & 24\% & 19\% \\
    \texttt{creator} & Core & Short-text & Dataset creator(s) & 100\% & 100\% \\
    \texttt{publisher} & Core & Constrained & Publishing venue/org & 80\% & 74\% \\
    \texttt{datePublished} & Core & Constrained & Publication date & 100\% & 89\% \\
    \texttt{inLanguage} & Core & Constrained & Content language(s) & 81\% & 93\% \\
    \texttt{citeAs} & Core & Short-text & Citation format & 99\% & 61\% \\
    \texttt{isLiveDataset} & Core & Constrained & Actively updated? & 100\% & 8\% \\
    \midrule
    \texttt{dataCollection} & RAI & LLM Judge & Collection methodology & 100\% & 99\% \\
    \texttt{dataCollectionType} & RAI & LLM Judge & Collection type (controlled vocab) & 99\% & 99\% \\
    \texttt{dataCollectionMissingData} & RAI & LLM Judge & Missing data handling & 13\% & 5\% \\
    \texttt{dataCollectionRawData} & RAI & LLM Judge & Raw source data description & 100\% & 99\% \\
    \texttt{dataCollectionTimeframe} & RAI & LLM Judge & Collection period & 26\% & 21\% \\
    \texttt{dataImputationProtocol} & RAI & LLM Judge & Imputation methods & 4\% & 2\% \\
    \texttt{dataManipulationProtocol} & RAI & LLM Judge & Post-processing modifications & 69\% & 62\% \\
    \texttt{dataPreprocessingProtocol} & RAI & LLM Judge & Data cleaning/formatting steps & 94\% & 96\% \\
    \texttt{dataAnnotationProtocol} & RAI & LLM Judge & Annotation methodology & 93\% & 85\% \\
    \texttt{dataAnnotationPlatform} & RAI & LLM Judge & Annotation platform & 34\% & 34\% \\
    \texttt{dataAnnotationAnalysis} & RAI & LLM Judge & Annotation quality analysis & 67\% & 43\% \\
    \texttt{annotationsPerItem} & RAI & LLM Judge & Labels per data item & 57\% & 41\% \\
    \texttt{annotatorDemographics} & RAI & LLM Judge & Annotator demographics & 43\% & 30\% \\
    \texttt{machineAnnotationTools} & RAI & LLM Judge & ML annotation tools & 72\% & 79\% \\
    \texttt{dataReleaseMaintenancePlan} & RAI & LLM Judge & Versioning/maintenance plan & 51\% & 35\% \\
    \texttt{personalSensitiveInformation} & RAI & LLM Judge & Sensitive attributes & 30\% & 26\% \\
    \texttt{dataSocialImpact} & RAI & LLM Judge & Social implications & 66\% & 42\% \\
    \texttt{dataBiases} & RAI & LLM Judge & Known biases & 74\% & 49\% \\
    \texttt{dataLimitations} & RAI & LLM Judge & Known limitations & 95\% & 90\% \\
    \texttt{dataUseCases} & RAI & LLM Judge & Intended use cases & 100\% & 100\% \\
    \midrule
    \textbf{Average} & & & & \textbf{72.3\%} & \textbf{62.3\%} \\
    \bottomrule
    \end{tabular}
\end{table*}

\subsection{Annotation Pipeline Phases}
\label{app:annotation-phases}

The pre-fill extraction ran in mid-January~2026: a single-pass batch over all 602 papers using Claude Sonnet 4.5 populated the 30-field schema for every paper at temperature~0 with the canonical prompt (Appendix~\ref{app:prompts}).

The 22-rater pool was assembled across seven phases between March and April 2026, with each phase motivated by a coverage gap exposed in the previous one. Cumulative coverage of the $102 \times 30 = 3{,}060$-cell grid grew monotonically from the calibration pilot to the orphan top-up that brought every cell to at least three distinct annotators; per-phase rating counts and design rationale follow.

\textbf{Annotator Selection and Calibration.} We received 27 annotator registrations, of which 25 reported regular hands-on experience with ML datasets and 24 reported prior experience reading dataset papers. A total of 21 annotators participated in a calibration pilot (18 of whom completed the 15-cell sample, with the remaining three submitting partial responses), and 22 distinct annotators contributed to the final dataset across all phases.
During calibration, the lead author manually graded each submission against the cited paper passage along three axes: (i) rubric adherence, (ii) quality and clarity of free-text notes and corrections, and (iii) accuracy relative to the source. Based on this evaluation, the 21 calibrated annotators were grouped into three quality tiers: Tier A (5 annotators: 1 best, 1 domain expert, 1 strict, 2 standard), Tier B (7 annotators), and Tier C (9 annotators). One additional annotator joined at the redistribution phase without calibration signals and was assigned a conservative workload.

\textbf{Task Assignment.} Task assignment followed two principles. First, quality tier determined difficulty: Tier A annotators were assigned the most challenging fields (typically 3 hardest, 2 hard, and 1 medium per rotation), Tier B handled predominantly hard fields, and Tier C focused on medium-to-easy fields. Second, workload tier determined volume, with heavy-load annotators completing approximately 600 ratings and light-load annotators approximately 200.
During annotation, the lead author continuously spot-checked submissions and reassigned tasks when ratings did not match the source paper. Annotator engagement was monitored throughout, with dropout risk categorized as LOW (19 annotators), MEDIUM (2), or HIGH (2). Two HIGH-risk annotators disengaged before completing their assignments; their remaining tasks were reassigned during redistribution phases.
We release all individual ratings with pseudonymous annotator IDs, a pseudonymized annotator table (role, experience, and workload), and the blank recruitment form.

\textbf{Recruitment and Compensation.} We recruited 22 annotators across 18 affiliations through an open research call. The group comprised four faculty members, two postdoctoral researchers, three researchers, seven industry or independent practitioners, three PhD students, and three MSc students. Participation was voluntary and unpaid; eight annotators are co-authors of this paper. Annotators contributed a median of 461 ratings each (range: 12--1{,}123). The public registration form described the study and collected contact details, affiliation, role, self-reported experience with ML datasets and dataset papers, and availability. The annotation instructions disclosed that the pre-filled values had been generated by an AI model.

\textbf{Human Annotation Pipeline.} Collection ran in four functional stages, totalling seven sequential phases, between March and April~2026: (i)~a 15-cell calibration pilot fixed the rubric and graded each annotator into a quality tier (Phase 1); (ii)~a full-grid pass over the $102 \times 30$ cells under the locked rubric (Phase 2); (iii)~redistribution and clean-up sweeps (Phases 3--7) brought every cell to at least three distinct annotators; and (iv)~senior adjudication resolved the 140 cells that did not reach majority. We aggregated 9{,}595 ratings post-deduplication; final gold values were derived via majority vote per cell.

\begin{figure}[!ht]
  \centering
  \includegraphics[width=0.8\linewidth]{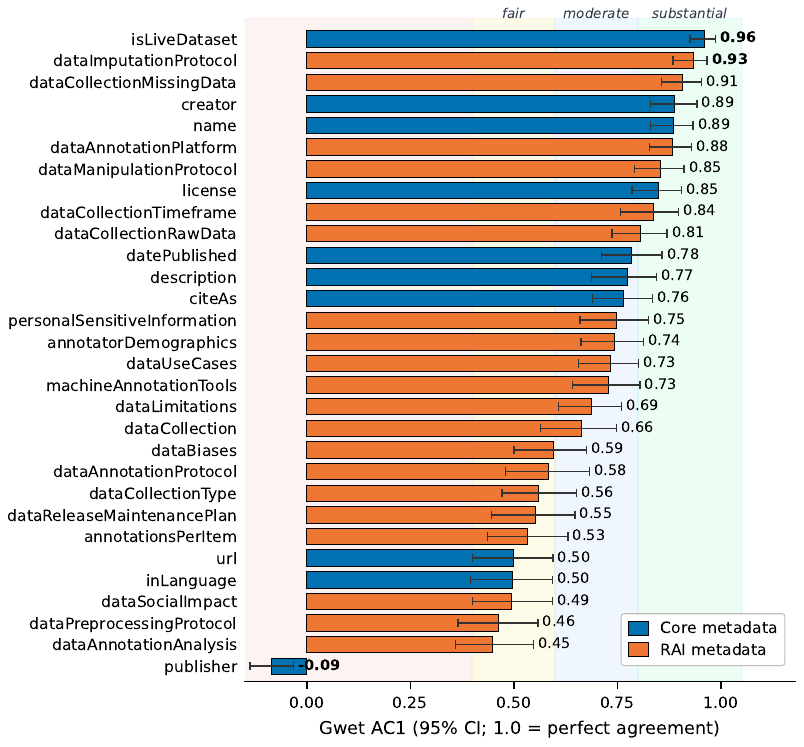}
  \caption{Annotation reliability for each metadata field. Bars show Gwet's AC$_1$ with 95\% bootstrap confidence intervals over 3{,}060 cells (102 papers $\times$ 30 fields) derived from 9{,}595 ratings from 22 annotators. AC$_1$ is reported alongside Krippendorff's $\alpha$ to account for skewed label distributions, where $\alpha$ can underestimate agreement. Background colors indicate Landis--Koch thresholds (fair $\geq 0.4$, moderate $\geq 0.6$, substantial $\geq 0.8$). Agreement is high overall (95.0\% majority consensus) but varies by field, with constrained fields showing higher agreement than open-ended fields.}
  \label{fig:annotation_results}
\end{figure}

\textbf{Phase 1 (early March 2026, calibration).} 18 annotators rated a 15-cell sample using a binary \texttt{TRUE}/\texttt{FALSE} rubric (267 ratings). The pilot exposed that the binary rubric could not capture partial correctness, motivating the three-level rubric used from Phase 2 onwards.

\textbf{Phase 2 (March--early April, full grid).} The three-level rubric (\textit{Correct} / \textit{Partially Correct} / \textit{Not Correct}) and six-category failure-mode taxonomy were introduced; annotators ran a full sweep over the $102 \times 30$ grid (6{,}658 ratings). Coverage proved uneven, leaving some cells with fewer than three distinct raters.

\textbf{Phase 3 (mid-April, redistribution).} Cells were redistributed across the available annotator pool to balance load (1{,}802 ratings).

\textbf{Phase 4 (mid-to-late April, three-rater target).} Targeted assignments brought most cells to the $\geq 3$-rater target (1{,}501 ratings); 208 cells remained under target after this phase.

\textbf{Phases 5--7 (late April, clean-up).} Phase 5 was a focused sweep that filled 204 of the 208 cells; it uncovered 28 additional cells that Phase 6 addressed (27 of 28 via a single rater). Phase 7 was the orphan top-up: the lead author served as the third rater for the remaining 8 cells, bringing every one of the 3{,}060 cells to at least three distinct annotators.

\textbf{Senior Adjudication.} A senior author resolved the 140 cells that did not reach majority (\texttt{tie\_split}) on top of the Phase 1--7 ratings. A separate batch of 14 cells where the lead-author orphan top-up arrived too late
for majority aggregation was resolved by the lead author and is logged under \texttt{gold\_method = orphan\_topup} (excluded from the 4.6\% adjudication share reported in the main text).

\subsection{Annotation Reliability}
\label{app:annotation_reliability}

We analyse inter-annotator agreement across all metadata fields to assess the reliability and difficulty of the annotation task (Fig.~\ref{fig:annotation_results}).

\subsection{Robustness to the Seed Model}
\label{app:seed-robustness}

Because the gold labels began as Claude Sonnet~4.5 pre-fills, a natural concern is that the benchmark rewards resemblance to Sonnet~4.5 rather than correct extraction. If that were true, changing the model that produced the pre-fills should reshuffle the ranking, and the cells where annotators replaced the pre-fill should order the systems differently. Neither happens: the checks below indicate that the ranking reflects extraction quality, while the seed model itself, and to a lesser degree its family, is favoured by the gold it seeded.

\paragraph{Re-annotation from a Different Seed Model} We re-annotated 10 test papers from scratch, starting from GPT-5.4 pre-fills instead of Sonnet~4.5 pre-fills and following the original protocol: three raters per cell, the same three-level rubric and majority voting. This gives a second gold set of 290 cells that owes nothing to Sonnet~4.5; Table~\ref{tab:seed-pilot} reports every system under both gold sets. Leaving out the two seed models, the rankings under the two gold sets correlate at Spearman $\rho = 0.903$, and the remaining systems move by only 0.026 on average. The seed model itself, however, gains a large advantage: GPT-5.4 rises from 0.673 to 0.930 when scored against gold it seeded, while Sonnet~4.5 falls from 0.824 to 0.634 once the gold is no longer its own, which is why we exclude the seed model from the ranking in Table~\ref{tab:test88-headline}. A smaller part of the effect reaches other models from the seed's family, such as Opus~4.7 and GPT-5.4 Mini, so comparisons across model families call for some care. Sonnet~4.6, by contrast, changes by only 0.014 and is the strongest non-seed system under the new gold. The main architectural finding also holds, as single-pass extraction still scores higher than the agentic architectures for every backbone. Because the re-annotation covers only 10 papers, it supports the overall ranking but not conclusions about individual fields.

\paragraph{Human-Rewritten Cells and Paraphrased Gold} Two further checks point the same way. In 376 of the 3{,}060 gold cells, or 12.3\%, annotators replaced the pre-fill with their own value, so resembling Sonnet~4.5 cannot help a system there; ranking the systems on these cells alone gives almost the same order as the full benchmark, with a Spearman correlation of 0.882. And rewriting every free-text gold value on 30 papers with two paraphrasing models from different families, which changes the wording but keeps the meaning, leaves the ranking unchanged, with a Spearman correlation of 0.98.

\begin{table}[!ht]
    \centering
    \scriptsize
    \caption{Re-annotation pilot on 10 test papers, laid out as Table~\ref{tab:test88-headline}. Core, RAI and the first composite are scored against the GPT-5.4-seeded gold; the last column is the composite against the current gold on the same papers. Systems are ranked by the pilot-gold composite within each architecture group, and $^{*}$ marks Anthropic-family systems. The two seed models, GPT-5.4 for the pilot gold and Sonnet~4.5 for the current gold, are inflated against the gold they seeded and are listed separately without a rank. The two Gemini-based Locator-Extractor runs are omitted because their pilot scores include failed API calls that were later re-run.}
    \label{tab:seed-pilot}
    \resizebox{\textwidth}{!}{%
    \begin{tabular}{llccccc}
    \toprule
    Rank & System & Architecture & Core & RAI & Composite, pilot gold [95\% CI] & Composite, current gold \\
    \midrule
    1 & \textbf{Claude Sonnet 4.6$^{*}$} & Single-Pass & \textbf{0.712} & \textbf{0.666} & \textbf{0.681}\, [0.627, 0.749] & 0.695 \\
    2 & GPT-5.4 Mini & Single-Pass & 0.702 & 0.665 & 0.678\, [0.613, 0.733] & 0.608 \\
    3 & Claude Opus 4.7$^{*}$ & Single-Pass & 0.695 & 0.619 & 0.645\, [0.585, 0.704] & 0.708 \\
    4 & GLM-5.1 & Single-Pass & 0.671 & 0.511 & 0.566\, [0.503, 0.632] & 0.581 \\
    5 & Gemini 3.1 Pro Preview & Single-Pass & 0.622 & 0.524 & 0.559\, [0.508, 0.605] & 0.546 \\
    6 & Qwen 3.6 35B-A3B & Single-Pass & 0.619 & 0.516 & 0.553\, [0.499, 0.617] & 0.591 \\
    7 & DeepSeek V3.2 & Single-Pass & 0.662 & 0.480 & 0.545\, [0.470, 0.591] & 0.560 \\
    8 & Gemini 2.5 Flash & Single-Pass & 0.644 & 0.492 & 0.543\, [0.466, 0.633] & 0.516 \\
    9 & Mistral Small 4 & Single-Pass & 0.630 & 0.451 & 0.511\, [0.473, 0.569] & 0.508 \\
    10 & Llama 4 Scout 17B & Single-Pass & 0.506 & 0.310 & 0.375\, [0.319, 0.466] & 0.383 \\
    \midrule
    1 & ReAct (GPT-5.4) & ReAct & 0.598 & 0.521 & 0.548\, [0.465, 0.614] & 0.607 \\
    2 & ReAct (Sonnet 4.6)$^{*}$ & ReAct & 0.526 & 0.506 & 0.513\, [0.411, 0.632] & 0.574 \\
    3 & ReAct (Gemini 3.1 Pro) & ReAct & 0.556 & 0.357 & 0.428\, [0.327, 0.515] & 0.507 \\
    \midrule
    1 & Parallel Specialists (GPT-5.4) & Parallel Specialists & \textbf{0.689} & \textbf{0.632} & \textbf{0.652}\, [0.597, 0.703] & 0.579 \\
    2 & Parallel Specialists (Sonnet 4.6)$^{*}$ & Parallel Specialists & 0.625 & 0.617 & 0.620\, [0.554, 0.669] & 0.633 \\
    3 & Parallel Specialists (Gemini 3.1 Pro) & Parallel Specialists & 0.519 & 0.481 & 0.494\, [0.442, 0.565] & 0.516 \\
    \midrule
    1 & Triage + Critique (GPT-5.4) & Triage + Critique & 0.582 & 0.525 & 0.545\, [0.489, 0.622] & 0.506 \\
    2 & Triage + Critique (Sonnet 4.6)$^{*}$ & Triage + Critique & 0.561 & 0.498 & 0.520\, [0.475, 0.582] & 0.562 \\
    3 & Triage + Critique (Gemini 3.1 Pro) & Triage + Critique & 0.475 & 0.282 & 0.351\, [0.297, 0.412] & 0.337 \\
    \midrule
    1 & Locator-Extractor (GPT-5.4) & Locator-Extractor & 0.547 & 0.483 & 0.506\, [0.451, 0.571] & 0.442 \\
    2 & Locator-Extractor (Sonnet 4.6)$^{*}$ & Locator-Extractor & 0.532 & 0.427 & 0.462\, [0.409, 0.538] & 0.480 \\
    \midrule
    \multicolumn{7}{l}{\textit{Unranked: seed models of the two gold sets}} \\
    -- & \textit{GPT-5.4} & \textit{Single-Pass} & \textit{0.933} & \textit{0.928} & \textit{0.930}\, \textit{[0.886, 0.965]} & \textit{0.673} \\
    -- & \textit{Claude Sonnet 4.5$^{*}$} & \textit{Single-Pass} & \textit{0.668} & \textit{0.615} & \textit{0.634}\, \textit{[0.566, 0.682]} & \textit{0.824} \\
    \bottomrule
    \end{tabular}%
    }
\end{table}

\FloatBarrier
\section{Synthesis and Implications}
\label{sec:discussion}

\textbf{Deployment as Author-in-the-loop Verification.} The leader composite of 0.709 is not enough for unattended metadata production, but it is a useful starting point for an author-facing verification tool. We position the released \cm system as a draft-then-review workflow: the system fills the 30 fields from the manuscript text and flags low-confidence fields back to the authors at submission time. Confidence can be set per-field from the per-field accuracy and IAA that the benchmark already provides, so authors review the fields where the model is empirically weakest (\texttt{sc:publisher}, \texttt{rai:dataCollectionType}, \texttt{rai:dataAnnotationAnalysis}; Section~\ref{sec:error-analysis}) rather than every field uniformly. Centralized deployment at journals and community repositories, combined with deterministic tooling for the structurally-extractable fields\footnote{\url{https://github.com/MIT-LCP/croissant-baker}}, amortizes the per-paper cost across many users.

\textbf{Ethical and Responsible AI Considerations.} The benchmark surfaces a systematic bias toward non-null outputs for undocumented fields (Section~\ref{sec:error-analysis}). For Responsible-AI fields (\texttt{rai:dataBiases}, \texttt{rai:dataSocialImpact}, \texttt{rai:personalSensitiveInformation}), interpretive leaps in place of nulls misrepresent what the dataset paper actually claims, which is exactly the failure mode policy and legal review of dataset documentation must avoid. We therefore release the benchmark with per-field accuracy and IAA so downstream users can decide which fields to accept automatically and which to require a human check; fields with low inter-annotator agreement should require explicit author sign-off before any extracted value is published as dataset metadata.

\section{Related Work (Extended)}
\label{app:related-extended}

This appendix provides an extended discussion of the related work summarised in Section~\ref{sec:intro}. We first expand on current works ranging from data documentation frameworks to machine-readable standards, then we dive into automated metadata and documentation generation methods, culminating in a broader literature on LLM-based structured extraction and evaluation that has inspired the design of our extraction systems.

\subsection{From Documentation Frameworks to Machine-Readable Standards}

The recognition that ML datasets require systematic documentation has produced a rich landscape of frameworks over the past decade. \citet{gebru2021datasheets} proposed \emph{Datasheets for Datasets}, drawing an analogy to electronics datasheets with 57 questions spanning motivation, composition, collection, and maintenance. Parallel efforts introduced \emph{Data Statements}~\citep{bender2018data} for NLP, \emph{Model Cards}~\citep{mitchell2019model} for disaggregated model evaluation, \emph{Data Cards}~\citep{pushkarna2022data} as participatory documentation artifacts, and \emph{Data Nutrition Labels}~\cite{holland2018dataset} for quantitative quality assessment. These frameworks reflect the data-centric AI movement's emphasis on treating data quality as a first-class concern~\citep{zha2025data}, responding to evidence that 92\% of AI practitioners experience \emph{data cascades} caused by inadequate data practices~\citep{sambasivan2021everyone}. Recent work moves dataset documentation toward scalable evaluation, but large-scale analyses still reveal persistent gaps in dataset cards~\citep{yang2024navigating}. Rubric-based frameworks aim to standardize the evaluation of dataset development and documentation practices~\citep{bhardwaj2024machine}, while more recent work supports automated quality metrics and accountability for dataset documentation~\citep{winata2025datasheets}. The EU AI Act further raises the stakes by mandating structured data governance and training data disclosure (Article~11,~\citealp{euaiact2024art11}). However, a major shortcoming of these frameworks is that they produce natural-language documents that require substantial manual effort and are difficult to standardize or reuse. Therefore, adoption remains limited: the analysis of NeurIPS submissions reveals that only 40\% of dataset papers address ethical concerns~\cite{wu2024systematic}, while the responsible AI sections of Hugging Face dataset cards are completed less than 2\% of the time~\cite{yang2024navigating}. 

Croissant~\citep{jain2024standardized} unified these threads by creating an ML-specific metadata format built on Schema.org with four layers: dataset metadata, resources, record structure, and ML semantics. Its RAI extension~\citep{jain2024standardized} adds 20 responsible AI attributes synthesized from Datasheets, Data Cards, Data Statements, and CrowdWorkSheets~\citep{diaz2022crowdworksheets}. Croissant has achieved remarkable adoption: over 700,000 datasets carry Croissant metadata~\citep{mlcommons2025croissant} across Hugging Face~\citep{lhoest2021datasets}, Kaggle, OpenML~\citep{vanschoren2014openml}, and Dataverse, and Croissant is now required for NeurIPS dataset submissions~\citep{neurips2026edfaq}.

Despite this widespread adoption, a critical gap remains in metadata coverage. Platform-generated Croissant metadata covers only structural fields inferred from data files.\footnote{Hugging Face Croissant generation: \url{https://huggingface.co/docs/dataset-viewer/croissant}} The contextual and responsible AI fields, describing data provenance, collection methodology, annotator demographics, intended use, and known biases, exist only in the scientific papers that introduce these datasets. Our extraction systems automate the extraction of these fields from papers, addressing the fundamental bottleneck preventing RAI metadata from reaching the ecosystem.

\subsection{Automated Metadata and Documentation Generation}

Several systems have explored the automation of dataset and model documentation. CardGen~\citep{liu2024cardgen} uses retrieval-augmented generation to produce model and data cards from papers and GitHub repositories, contributing CardBench with 4,800 model cards and 1,400 data cards, though its output is free-text narrative rather than structured metadata. Datadoc Analyzer~\citep{giner2023datadoc} targets data papers beyond the ML community, extracting dimensions closely aligned with those later formalized in the Croissant RAI extension, but producing outputs tied to its own schema. MOLE~\citep{alyafeai2025mole} extracts over 30 metadata attributes from papers using LLMs, with Gemini 2.5 Pro achieving 67.42\%; MeXtract~\citep{alyafeai2025mextract} extends it with fine-tuned sub-3B models via knowledge distillation. MOLE and MeXtract are built around the Masader schema, which shares some structural and RAI dimensions with Croissant but neither adopts the Croissant-RAI attribute set nor produces Croissant-conformant output. Retargeting would require redefining the attribute set, retraining or re-prompting models to match RAI-specific and Croissant extraction patterns, and rebuilding the evaluation data; effectively the contribution of \cm. More fundamentally, none of these systems produces output that conforms to the standard the ecosystem has converged on, which is a prerequisite for metadata to flow through the 700,000+ datasets and platforms that already consume Croissant.

On the platform side, Hugging Face automatically generates Croissant JSON-LD for datasets in supported formats, inferring structural metadata (column names, types, distributions) from data files. OpenML~\citep{vanschoren2014openml} and Kaggle provide similar auto-generation capabilities. These platform tools are complementary to \cm: they populate structural and statistical fields that \cm does not target, while \cm extracts the contextual and RAI fields that cannot be inferred from data alone.

Benchmarks for evaluating dataset metadata extraction remain scarce and narrowly scoped. Masader~\citep{alyafeai2025mole} provides gold annotations for Arabic NLP datasets against its own schema, and CardBench~\citep{liu2024cardgen} evaluates free-text card generation rather than structured field extraction. Neither targets Croissant nor its RAI extension, and neither captures the long-tail contextual fields (collection methodology, annotator demographics, known biases, intended use) that responsible AI documentation requires. The benchmark released with this paper is, to our knowledge, the first to pair scientific papers with Croissant-RAI-conformant gold annotations, enabling direct, standard-aligned evaluation of extraction systems and providing a foundation for future work.

\begin{table}[t]
    \caption{Comparison of automated dataset documentation systems.}
    \label{tab:comparison}
    \centering
    \small
    \begin{tabular}{lccccc}
    \toprule
    & \textbf{CardGen} & \textbf{MOLE} & \textbf{HF Auto} & \textbf{DataDoc} & \textbf{\cm} \\
    \midrule
    Input source & Paper + repo & Paper & Data files & Paper & Paper \\
    Output format & Free text & Structured & JSON-LD & Free text & JSON-LD \\
    Target schema & None & Masader & Croissant & None & Croissant 1.1 \\
    Core fields & Partial & 30+ & \checkmark & \texttimes & \checkmark~(10) \\
    RAI fields & Partial & \texttimes & \texttimes & Partial & \checkmark~(20) \\
    Croissant-conformant & \texttimes & \texttimes & \texttimes & \texttimes & \checkmark \\
    Human evaluation & \texttimes & \texttimes & \texttimes & \checkmark & \checkmark \\
    Benchmark size & 1.4K cards & 126 papers & N/A & N/A & 602 datasets \\
    \bottomrule
    \end{tabular}
\end{table}

\subsection{LLM-Based Structured Extraction and Evaluation}

The broader literature on LLM-based information extraction provides \cm's technical foundation. Early neural approaches to SciIE~\citep{luan2018multitask, jain2020scirex} relied on supervised models trained on domain-specific corpora. Recent work has demonstrated that LLMs substantially outperform these systems. \citet{dagdelen2024structured} showed that fine-tuning with as few as 20 examples enables accurate JSON-structured extraction from scientific text. ~\cite{polak2024extracting} achieved approximately 90\% precision and recall on materials data extraction using conversational prompt engineering with iterative verification. For guaranteeing schema-conformant output, constrained decoding~\citep{willard2023efficient} enforces structural validity during generation, while specialized models like NuExtract~\citep{numind2024nuextract} are purpose-built for template-guided JSON extraction. GoLLIE~\citep{sainz2024gollie} demonstrates that fine-tuning models to follow annotation guidelines substantially improves zero-shot IE, motivating our use of detailed per-field extraction guides in the CroissantMiner prompt. Closer to our setting, LMDX~\citep{perot2024lmdx} extends document IE with field localization, while Data Gatherer~\citep{marini2025data} extracts dataset references from scientific literature. Recent work advances schema-guided extraction, with PARSE~\citep{shrimal2025parse} optimizing JSON schemas and SPT~\citep{liang2025schema} treating schemas as callable tools for adaptive extraction.

Evaluating extraction quality at scale requires methods beyond exact match. \cite{zheng2023judging} demonstrated that LLM judges achieve over 80\% agreement with human preferences, comparable to inter-annotator agreement. G-Eval~\citep{liu2023geval} showed that chain-of-thought prompting improves LLM evaluation alignment with human judgments. Self-consistency~\citep{wang2023selfconsistency}, sampling multiple reasoning paths and selecting the most frequent answer, provides an orthogonal strategy for improving extraction reliability without additional supervision. Recent benchmarks strengthen structured extraction evaluation, with ExtractBench~\citep{ferguson2026extractbench} introducing fine-grained key-level metrics, LLMStructBench~\citep{tenckhoff2026llmstructbench} evaluating schema-constrained structured outputs, and StructEval~\citep{yang2026structeval} broadly assessing LLMs' structural generation capabilities.

Table~\ref{tab:comparison} positions \cm relative to existing systems. Unlike CardGen, which produces narrative documentation, and MOLE, which targets the Masader schema, our systems extract all 30 Croissant 1.1 fields as machine-readable JSON-LD that integrates directly with the ecosystem of 700,000+ Croissant-enabled datasets.

\begin{figure}[!ht]
    \centering
    \begin{tcolorbox}[
      enhanced,
      colback=white,
      colframe=black,
      title={Croissant Metadata Extraction Prompt},
      fonttitle=\bfseries,
      fontupper=\small,
    ]
    You are an expert researcher specializing in ``Responsible AI'' and
    metadata extraction for Machine Learning datasets.
    
    \medskip
    \textbf{Your Task:} Extract structured metadata from the provided
    academic paper according to the MLCommons Croissant RAI schema
    described below.
    
    \medskip
    \textbf{Extraction Guidelines:}
    \begin{enumerate}[leftmargin=1.5em, itemsep=0pt, topsep=2pt]
      \item \textit{Accuracy First:} Only extract information explicitly stated in the paper.
      \item \textit{No Guessing:} If information is not found, return \texttt{null}. Do not fill fields with your own background information or assumptions that are not explicitly given in the paper.
      \item \textit{Formatting Rules:} Return ONLY valid JSON\@. Use \texttt{null} for missing fields. No markdown formatting.
      \item \textit{RAI Field-Specific Information:} Use the following official definitions to guide your extraction. Use them as matching criteria when scanning the text.
    \end{enumerate}
    
    \medskip
    {[Per-field definitions for all 20 RAI fields and 10 Core fields
    follow here, drawn verbatim from the Croissant 1.1 specification.
    For brevity the full enumeration is omitted; the source of truth
    is \texttt{croissantminer/config.py::SYSTEM\_PROMPT} in the released code.]}
    
    \medskip
    The user prompt template appends the cleaned paper text and the
    exact 30-field schema (\texttt{sc:name}, \texttt{sc:description}, \texttt{sc:url}, \texttt{sc:license},
    \texttt{sc:creator}, \texttt{sc:publisher}, \texttt{sc:datePublished}, \texttt{sc:inLanguage},
    \texttt{cr:citeAs}, \texttt{cr:isLiveDataset}, plus the 20 \texttt{rai:} fields) and asks for
    a single JSON object that conforms to that schema.
    \end{tcolorbox}
    \caption{Prompt used to extract Croissant metadata fields.}
    \label{fig:metadata_extraction_prompt}
\end{figure}

\section{Prompt Design and Extraction Details}
\label{app:prompts}

This appendix reproduces the two prompts that drive the released pipeline: the canonical extraction system prompt used by all single-pass and agentic-specialist calls (Section~\ref{sec:prompt-design}) and the Tier~2 LLM-judge prompt used by GLM-5 to score the prose RAI fields (Section~\ref{sec:experimental-setup}). Both are reproduced verbatim; the prompt SHA-256 prefix \texttt{1e1cfdd99246bbf5} is stamped into every released extraction's \texttt{\_meta} block for provenance.

\subsection{Canonical Extraction Prompt}
\label{app:prompts-extraction}

The system prompt (Figure~\ref{fig:metadata_extraction_prompt}) is structured in four blocks: a role framing that casts the model as a metadata-extraction expert, accuracy-first guidelines (no inference from background knowledge, return \texttt{null} when not stated), per-field definitions drawn from the Croissant 1.1 specification, and the canonical schema. The full text is approximately 2{,}100 words and is reproduced below (formatted for readability; the released form is plain text).

\subsection{LLM Judge Prompt (Tier-2)}
\label{app:prompts-judge}

\begin{figure}[t]
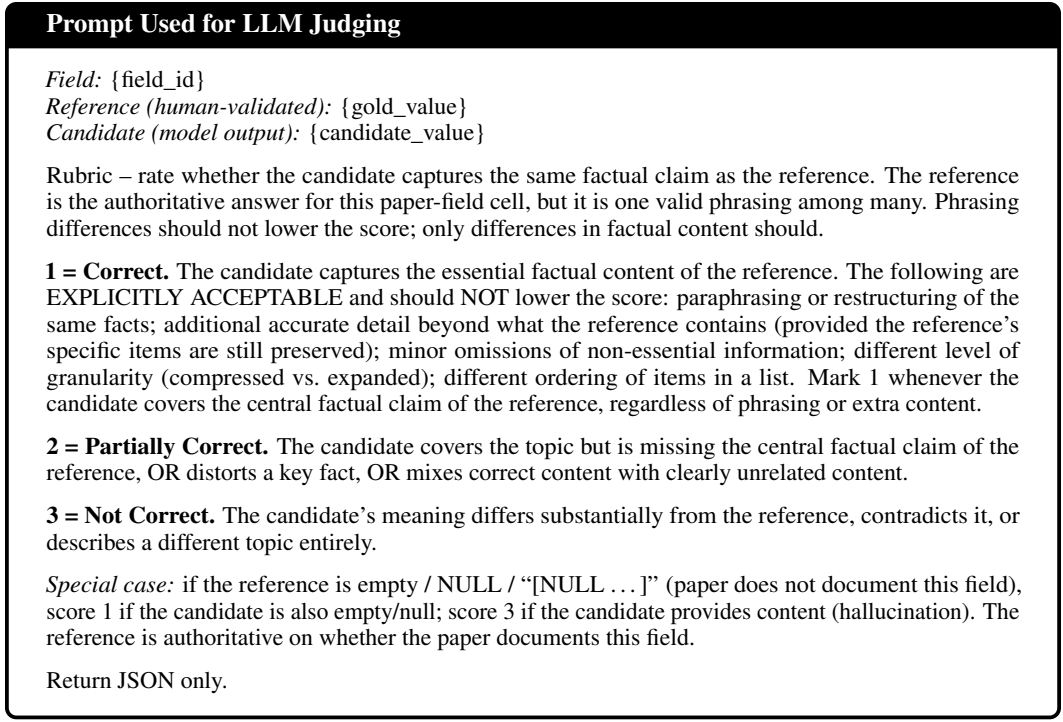

    \centering
    \begin{tcolorbox}[
      enhanced,
      colback=white,
      colframe=black,
      title={Prompt Used for LLM Judging},
      fonttitle=\bfseries,
      fontupper=\small,
    ]
    \textit{Field:} \{field\_id\}\\
    \textit{Reference (human-validated):} \{gold\_value\}\\
    \textit{Candidate (model output):} \{candidate\_value\}
    
    \medskip
    Rubric -- rate whether the candidate captures the same factual claim
    as the reference. The reference is the authoritative answer for this
    paper-field cell, but it is one valid phrasing among many. Phrasing
    differences should not lower the score; only differences in factual
    content should.
    
    \medskip
    \textbf{1 = Correct.} The candidate captures the essential factual
    content of the reference. The following are EXPLICITLY ACCEPTABLE and
    should NOT lower the score: paraphrasing or restructuring of the same
    facts; additional accurate detail beyond what the reference contains
    (provided the reference's specific items are still preserved); minor
    omissions of non-essential information; different level of granularity
    (compressed vs.\ expanded); different ordering of items in a list.
    Mark 1 whenever the candidate covers the central factual claim of the
    reference, regardless of phrasing or extra content.
    
    \medskip
    \textbf{2 = Partially Correct.} The candidate covers the topic but is
    missing the central factual claim of the reference, OR distorts a key
    fact, OR mixes correct content with clearly unrelated content.
    
    \medskip
    \textbf{3 = Not Correct.} The candidate's meaning differs substantially
    from the reference, contradicts it, or describes a different topic
    entirely.
    
    \medskip
    \textit{Special case:} if the reference is empty / NULL / ``[NULL \ldots]''
    (paper does not document this field), score 1 if the candidate is also
    empty/null; score 3 if the candidate provides content (hallucination).
    The reference is authoritative on whether the paper documents this
    field.
    
    \medskip
    Return JSON only.
    \end{tcolorbox}
    \caption{Prompt used for structured LLM-based judging (final rubric, used for all reported scores).}
    \label{fig:tier2_judge_prompt}
    \vspace{-10pt}
\end{figure}

GLM-5 receives one prompt per (paper, field) cell (Figure~\ref{fig:tier2_judge_prompt}). The reference (human-validated gold) is provided alongside the candidate extraction; the judge returns a structured JSON containing a 1--3 score and a free-text reason.

The judge is called via DeepInfra (\path{zai-org/GLM-5}) with
\path{temperature=0}, \path{max_completion_tokens=4000}, and 30 concurrent workers; failed calls are retried up to three times with exponential backoff. Scores are parsed by regex from the JSON response; cells with a parse failure are excluded from the per-system composite (this affected fewer than 1\% of cells in the released runs).

\textbf{Human Audit of the Deployed Judge.} Two authors independently rated 200 RAI cells using the three-point rubric. A third author assessed the 56 cells on which they disagreed and provided the deciding rating; we refer to the resulting labels as human consensus. The audit covers all 20 RAI fields, with 10 cells per field drawn from 18 systems, and is enriched for difficult cases. The judge achieves 71.5\% exact agreement with human consensus ($\kappa = 0.663$); agreement between the two initial raters is 72.0\% ($\kappa = 0.739$). Table~\ref{tab:judge-audit-perfield} reports exact agreement by field. Given the small sample per field, we report agreement counts rather than per-field $\kappa$. Of the 57 judge--consensus disagreements, 45 involve a more lenient judgment by the judge. Such judgments occur in 26.1\% of audited non-Anthropic cells and 13.8\% of Anthropic-family cells. Across the 18 audited systems, mean scores from the judge and human consensus have a Spearman rank correlation of $\rho = 0.746$ ($p < 0.001$). These results describe the audited cells and system versions, which do not exactly match the full lineup in Table~\ref{tab:test88-headline}.

\begin{table}[!ht]
    \centering
    \scriptsize
    \caption{Exact agreement between the deployed judge and human consensus on the 200-cell audit. Each entry gives the number of matching ratings out of 10 audited cells per RAI field.}
    \label{tab:judge-audit-perfield}
    \begin{tabular}{lr@{\hspace{2em}}lr}
    \toprule
    Field & Agree & Field & Agree \\
    \midrule
    annotatorDemographics & 10/10 & dataAnnotationPlatform & 7/10 \\
    annotationsPerItem & 9/10 & dataImputationProtocol & 6/10 \\
    dataCollectionTimeframe & 9/10 & dataCollection & 6/10 \\
    machineAnnotationTools & 9/10 & dataAnnotationAnalysis & 6/10 \\
    dataSocialImpact & 9/10 & personalSensitiveInformation & 6/10 \\
    dataBiases & 9/10 & dataCollectionRawData & 5/10 \\
    dataUseCases & 8/10 & dataLimitations & 5/10 \\
    dataPreprocessingProtocol & 8/10 & dataCollectionMissingData & 4/10 \\
    dataCollectionType & 8/10 & dataManipulationProtocol & 4/10 \\
    dataAnnotationProtocol & 8/10 & & \\
    dataReleaseMaintenancePlan & 7/10 & & \\
    \bottomrule
    \end{tabular}
\end{table}

\section{System Implementation Details}
\label{app:implementation}

\textbf{Pipeline.} Single-pass extractions are issued via the provider-native batch APIs (Anthropic Batch API for Claude Sonnet/Opus, OpenAI Batch API for GPT-5.4 family, Google Vertex batch for Gemini, vLLM 0.11.1 with \texttt{--max-model-len 200000} on institutional GPU nodes for self-hosted Llama~4 Scout, Qwen~3.6~35B-A3B, Mistral Small~4, DeepSeek~V3.2, GLM-5.1). All commercial-API calls fix \texttt{temperature=0} and \texttt{max\_tokens=4096}. Open-weight
runs use the same temperature; the vLLM serve is launched per backbone and the extraction script issues the 102 paper requests in series.

\textbf{Provenance.} Every extraction emits a \texttt{\_meta} block recording the model identifier (with version), the exact prompt SHA-256 prefix (\texttt{1e1cfdd99246bbf5} for the canonical extraction prompt), the wall time, the input and output token counts, and the parser version (\texttt{PyPDF2 3.0.1}). The \texttt{\_meta} block is included in every released JSON-LD record.

\textbf{Agentic Implementations.} The four agentic architectures share a provider abstraction (\texttt{scripts/\_agentic\_helpers.py}) that normalises Anthropic, OpenAI, Google, and DeepInfra into a single \texttt{call(model, prompt) -> response} interface. Each architecture is implemented as one top-level script (\texttt{scripts/agentic\_v2.py}, \texttt{scripts/agentic\_lev.py}, \texttt{croissantminer/react\_agent/}) with a configurable \texttt{--backbone} flag.

\textbf{Scoring.} Tier~1 (rule-based) scoring uses \texttt{evaluation/field\_metrics.py}: exact-match after canonicalisation for the seven enum-like fields
(\texttt{license}, \texttt{inLanguage}, \texttt{datePublished}, \texttt{publisher}, \texttt{url}, \texttt{name}, \texttt{isLiveDataset}), token~F1 for the three short-text fields (\texttt{description}, \texttt{creator}, \texttt{citeAs}). Tier~2 (LLM judge) calls GLM-5 via DeepInfra (\texttt{zai-org/GLM-5}) at 30 concurrent workers; the judge prompt is in Appendix~\ref{app:prompts}.

\textbf{Statistical Comparisons.}
\label{app:eval-methodology}
The composite weights all 30 fields equally and is reported with a 2{,}000-replicate paper-clustered bootstrap 95\%~CI~\citep{efron1979bootstrap}, where resampling is at the paper level (not the cell level) to respect within-paper correlation across fields. Pairwise system comparisons use three complementary tests on inner-joined cells: the paired bootstrap on per-cell composite deltas, the Wilcoxon signed-rank test~\citep{wilcoxon1945individual} on per-cell deltas, and McNemar's test~\citep{mcnemar1947note} on scores binarised at 0.5. All $p$-values are Benjamini--Hochberg-FDR adjusted~\citep{benjamini1995controlling} across the pairwise comparison budget. Cells where both gold and prediction are null are skipped (correct-null) so systems are not rewarded for trivially mass-emitting null; cells where gold is null but the prediction is non-null are scored zero (hallucination); cells where gold is non-null but prediction is null are also scored zero (missing).

\textbf{Evaluation Inference Cost.} From recorded token usage, we estimate that one pass of the 21 API-billed systems in Table~\ref{tab:test88-headline} over the 88 test papers costs approximately \$400 at standard list prices, or \$380 after applying the batch discount used for single-pass runs. These estimates cover extraction inference rather than the full development and evaluation process, and exclude unrecorded Gemini thinking tokens. The three self-hosted models used approximately 12 GPU hours on the institutional cluster, including smoke tests. Appendix~\ref{app:cost} reports per-system estimates and token-accounting limitations.

\section{Full Experimental Results}

\subsection{Per-Field Results}
\label{app:perfield}

Tables~\ref{tab:perfield-general} and~\ref{tab:perfield-rai} report per-field accuracy on the 88-paper test split for every system in the headline ranking (excluding Sonnet~4.5 per the gold-seed caveat of Section~\ref{sec:headline-ranking}). The core-field table covers the ten \texttt{sc:}/\texttt{cr:} fields scored by Tier~1 rule-based metrics (exact-match for the constrained fields, token-F1 for the short-text fields). The RAI table covers the twenty \texttt{rai:} fields scored by the Tier~2 GLM-5 LLM judge (1 = Correct, 0.5 = Partial, 0 = Wrong, mapped from the 1--3 ordinal rubric in Appendix~\ref{app:prompts-judge}).

\begin{table}[!ht]
    \centering
    \footnotesize
    \caption{Per-field accuracy on the 88-paper test split for the 10 core (sc:/cr:) fields. Scores in [0,1]; Tier 1 rule-based metric (exact-match for constrained, token-F1 for short-text). Higher is better. The last row gives the number of test papers (of 88) whose gold documents the field.}
    \label{tab:perfield-general}
    \resizebox{\textwidth}{!}{%
    \begin{tabular}{lrrrrrrrrrr}
    \toprule
    System & name & description & url & license & creator & publisher & datePublished & inLanguage & citeAs & isLiveDataset \\
    \midrule
    Claude Sonnet 4.6 & 0.98 & 0.70 & 0.85 & 0.42 & 0.74 & 0.62 & 0.90 & 0.70 & 0.76 & 0.85 \\
    Claude Opus 4.7 & 0.92 & 0.65 & 0.83 & 0.59 & 0.63 & 0.66 & 0.81 & 0.60 & 0.53 & 0.53 \\
    GPT-5.4 & 0.91 & 0.63 & 0.83 & 0.57 & 0.75 & 0.30 & 0.83 & 0.72 & 0.77 & 0.24 \\
    ReAct (4.6) & 0.94 & 0.64 & 0.84 & 0.27 & 0.81 & 0.62 & 0.88 & 0.73 & 0.72 & 0.90 \\
    PSpec (4.6) & 0.97 & 0.65 & 0.86 & 0.48 & 0.81 & 0.06 & 0.92 & 0.74 & 0.78 & 0.73 \\
    Qwen 3.6 35B-A3B & 0.81 & 0.61 & 0.80 & 0.39 & 0.78 & 0.59 & 0.90 & 0.72 & 0.45 & 0.94 \\
    GLM-5.1 & 0.97 & 0.65 & 0.85 & 0.39 & 0.76 & 0.48 & 0.91 & 0.64 & 0.62 & 0.50 \\
    Triage+Critique (4.6) & 0.97 & 0.67 & 0.85 & 0.13 & 0.81 & 0.45 & 0.94 & 0.71 & 0.38 & 0.83 \\
    Gemini 2.5 Flash & 0.76 & 0.54 & 0.83 & 0.35 & 0.77 & 0.48 & 0.89 & 0.64 & 0.50 & 0.41 \\
    GPT-5.4 Mini & 0.76 & 0.48 & 0.82 & 0.38 & 0.77 & 0.41 & 0.86 & 0.67 & 0.19 & 0.27 \\
    Gemini 3.1 Pro & 0.93 & 0.62 & 0.87 & 0.48 & 0.80 & 0.49 & 0.90 & 0.54 & 0.06 & 0.08 \\
    DeepSeek V3.2 & 0.83 & 0.62 & 0.80 & 0.46 & 0.77 & 0.45 & 0.85 & 0.66 & 0.55 & 0.17 \\
    Locator-Extr (4.6) & 0.97 & 0.66 & 0.70 & 0.14 & 0.77 & 0.67 & 0.88 & 0.73 & 0.06 & 0.85 \\
    Mistral Small 4 & 0.86 & 0.61 & 0.77 & 0.17 & 0.70 & 0.58 & 0.91 & 0.68 & 0.66 & 0.20 \\
    Llama 4 Scout & 0.75 & 0.37 & 0.78 & 0.35 & 0.77 & 0.54 & 0.57 & 0.50 & 0.26 & 0.16 \\
    \midrule
    Support ($n$) & 88 & 88 & 84 & 22 & 88 & 71 & 88 & 72 & 87 & 88 \\
    \bottomrule
    \end{tabular}%
    }
\end{table}

\begin{table}[!ht]
    \centering
    \footnotesize
    \caption{Per-field accuracy on the 88-paper test split for the 20 RAI (rai:) fields. Scores in [0,1] from the GLM-5 LLM judge (1=Correct, 0.5=Partial, 0=Wrong). Higher is better. The last row gives the number of test papers (of 88) whose gold documents the field; scores on fields with small support, such as (12) and (8), are unstable.\\
    \textbf{Field key:}
    (1)~annotationsPerItem,
    (2)~annotatorDemographics,
    (3)~dataAnnotationAnalysis,
    (4)~dataAnnotationPlatform,
    (5)~dataAnnotationProtocol,
    (6)~dataBiases,
    (7)~dataCollection,
    (8)~dataCollectionMissingData,
    (9)~dataCollectionRawData,
    (10)~dataCollectionTimeframe,
    (11)~dataCollectionType,
    (12)~dataImputationProtocol,
    (13)~dataLimitations,
    (14)~dataManipulationProtocol,
    (15)~dataPreprocessingProtocol,
    (16)~dataReleaseMaintenancePlan,
    (17)~dataSocialImpact,
    (18)~dataUseCases,
    (19)~machineAnnotationTools,
    (20)~personalSensitiveInformation.}
    \label{tab:perfield-rai}
    \resizebox{\textwidth}{!}{%
    \begin{tabular}{lrrrrrrrrrrrrrrrrrrrr}
    \toprule
    System & (1) & (2) & (3) & (4) & (5) & (6) & (7) & (8) & (9) & (10) & (11) & (12) & (13) & (14) & (15) & (16) & (17) & (18) & (19) & (20) \\
    \midrule
    Claude Sonnet 4.6 & 0.55 & 0.73 & 0.62 & 0.69 & 0.84 & 0.66 & 0.94 & 0.25 & 0.96 & 0.74 & 0.76 & 0.12 & 0.87 & 0.58 & 0.79 & 0.61 & 0.70 & 0.98 & 0.82 & 0.53 \\
    Claude Opus 4.7 & 0.54 & 0.76 & 0.63 & 0.75 & 0.82 & 0.65 & 0.93 & 0.50 & 0.90 & 0.83 & 0.71 & 0.33 & 0.81 & 0.57 & 0.75 & 0.66 & 0.75 & 0.95 & 0.81 & 0.55 \\
    GPT-5.4 & 0.56 & 0.64 & 0.64 & 0.68 & 0.84 & 0.67 & 0.98 & 0.40 & 0.93 & 0.79 & 0.73 & 0.00 & 0.84 & 0.60 & 0.71 & 0.71 & 0.67 & 0.99 & 0.65 & 0.40 \\
    ReAct (4.6) & 0.49 & 0.56 & 0.55 & 0.67 & 0.81 & 0.66 & 0.94 & 0.31 & 0.89 & 0.77 & 0.64 & 0.00 & 0.83 & 0.20 & 0.79 & 0.34 & 0.68 & 0.95 & 0.80 & 0.33 \\
    PSpec (4.6) & 0.34 & 0.46 & 0.59 & 0.68 & 0.67 & 0.59 & 0.96 & 0.22 & 0.95 & 0.68 & 0.73 & 0.00 & 0.85 & 0.47 & 0.79 & 0.63 & 0.76 & 0.97 & 0.74 & 0.33 \\
    Qwen 3.6 35B-A3B & 0.46 & 0.69 & 0.44 & 0.51 & 0.77 & 0.57 & 0.89 & 0.50 & 0.84 & 0.64 & 0.61 & 0.12 & 0.74 & 0.51 & 0.61 & 0.56 & 0.64 & 0.86 & 0.61 & 0.47 \\
    GLM-5.1 & 0.49 & 0.67 & 0.58 & 0.61 & 0.75 & 0.57 & 0.85 & 0.21 & 0.78 & 0.70 & 0.68 & 0.12 & 0.75 & 0.48 & 0.65 & 0.57 & 0.75 & 0.83 & 0.58 & 0.36 \\
    Triage+Critique (4.6) & 0.47 & 0.44 & 0.60 & 0.61 & 0.79 & 0.64 & 0.90 & 0.27 & 0.94 & 0.77 & 0.71 & 0.00 & 0.76 & 0.17 & 0.73 & 0.66 & 0.64 & 0.97 & 0.69 & 0.20 \\
    Gemini 2.5 Flash & 0.48 & 0.57 & 0.58 & 0.55 & 0.77 & 0.57 & 0.88 & 0.32 & 0.85 & 0.75 & 0.69 & 0.08 & 0.71 & 0.49 & 0.69 & 0.62 & 0.78 & 0.93 & 0.57 & 0.44 \\
    GPT-5.4 Mini & 0.55 & 0.57 & 0.54 & 0.66 & 0.74 & 0.53 & 0.90 & 0.16 & 0.88 & 0.69 & 0.70 & 0.17 & 0.73 & 0.58 & 0.68 & 0.62 & 0.64 & 0.90 & 0.70 & 0.31 \\
    Gemini 3.1 Pro & 0.46 & 0.62 & 0.46 & 0.78 & 0.66 & 0.42 & 0.88 & 0.61 & 0.83 & 0.76 & 0.60 & 0.14 & 0.59 & 0.41 & 0.60 & 0.45 & 0.70 & 0.91 & 0.57 & 0.48 \\
    DeepSeek V3.2 & 0.39 & 0.46 & 0.44 & 0.46 & 0.71 & 0.54 & 0.87 & 0.42 & 0.86 & 0.62 & 0.62 & 0.33 & 0.68 & 0.24 & 0.69 & 0.56 & 0.43 & 0.84 & 0.58 & 0.37 \\
    Locator-Extr (4.6) & 0.51 & 0.51 & 0.48 & 0.61 & 0.58 & 0.40 & 0.88 & 0.17 & 0.87 & 0.55 & 0.52 & 0.25 & 0.60 & 0.34 & 0.52 & 0.34 & 0.64 & 0.88 & 0.49 & 0.41 \\
    Mistral Small 4 & 0.32 & 0.39 & 0.43 & 0.37 & 0.68 & 0.57 & 0.94 & 0.07 & 0.84 & 0.26 & 0.63 & 0.04 & 0.68 & 0.19 & 0.64 & 0.42 & 0.63 & 0.90 & 0.52 & 0.12 \\
    Llama 4 Scout & 0.16 & 0.12 & 0.24 & 0.39 & 0.45 & 0.33 & 0.72 & 0.12 & 0.56 & 0.32 & 0.57 & 0.04 & 0.43 & 0.08 & 0.30 & 0.26 & 0.53 & 0.66 & 0.21 & 0.17 \\
    \midrule
    Support ($n$) & 53 & 38 & 61 & 33 & 82 & 68 & 88 & 11 & 88 & 24 & 88 & 2 & 85 & 63 & 84 & 43 & 61 & 88 & 66 & 26 \\
    \bottomrule
    \end{tabular}%
    }
\end{table}

\subsection{Silver split tracks the gold split}
\label{app:silver-vs-gold}

\begin{figure}[t]
  \centering
  \includegraphics[width=0.7\linewidth]{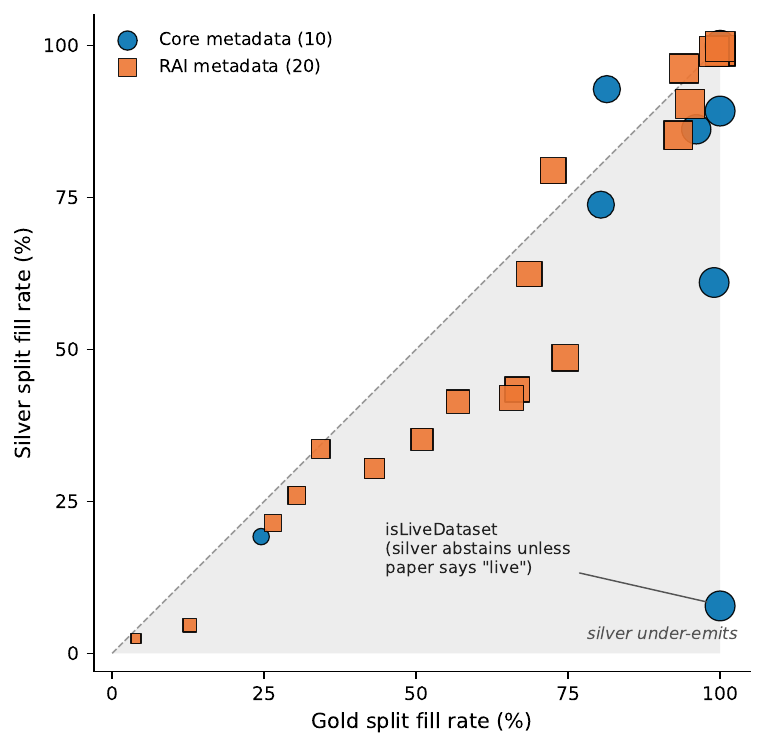}
  \caption{Silver-split fill rates track gold-split fill rates. Each point is one of the 30 Croissant fields; the $x$-coordinate is the fraction of the 102 human-validated gold papers in which the field is populated, and the $y$-coordinate is the same fraction in the 500 silver papers (Sonnet~4.5 reference). Pearson $r = 0.839$, Spearman $\rho = 0.823$ ($n = 30$, $p < 10^{-7}$). Core fields (blue circles) and RAI fields (orange squares) lie mostly on or below the diagonal: the silver extraction documents most fields less often than the human gold. The labelled outlier is \texttt{cr:isLiveDataset} (gold 100\%, silver 7.8\%): human annotators are instructed to decide for every paper, while the silver model only emits this field when the text explicitly references live updates.}
  \label{fig:silver_vs_gold}
\end{figure}

The 500-paper silver split is a Sonnet~4.5 single-pass extraction over the larger Hugging Face top-downloaded corpus, with no human verification. A natural concern is whether silver-only analyses generalise to the human-validated gold split. We compare per-field fill rates between the two splits in Figure~\ref{fig:silver_vs_gold}: the two are tightly correlated (Pearson $r = 0.839$, Spearman $\rho = 0.823$, $n = 30$, $p < 10^{-7}$), supporting the use of the silver split for scale-side studies (distillation, fine-tuning, corpus-level coverage trends). The largest disagreement is on \texttt{cr:isLiveDataset}, where annotators are instructed to decide for every paper while the model only emits the field when the text explicitly mentions live updates.

\subsection{PDF Parser Ablation}
\label{app:parser-ablation}

The pipeline converts PDFs to text with PyPDF2 (Section~\ref{sec:task-pipeline}), which linearises tables and figure text into plain lines and loses layout. To test whether this limits extraction quality, we re-ran single-pass extraction with PyMuPDF and with the layout-aware Docling, keeping the prompt, preprocessing, and scoring unchanged. We use Sonnet~4.6 and GPT-5.4 on 30 papers sampled at random (seed 42) from the test split. For Docling we use the default converter and export the full document: body text, page headers and footers, and text inside figures. Equations remain unconverted placeholders. All RAI cells in Table~\ref{tab:parser_ablation} were judged in a single session, so values should be compared within the table only.

Parser choice has little effect. With Docling's full export, all three parsers score within 0.01 overall for each model (Sonnet~4.6: 0.564--0.573; GPT-5.4: 0.532--0.541). RAI differences are at most 0.013 and point in different directions for the two models, and no RAI field improves consistently for both, including the table-heavy \texttt{rai:annotationsPerItem} and \texttt{rai:dataAnnotationAnalysis}. Docling's default Markdown export, by contrast, keeps only the body layer and discards page headers and footers, where papers usually print the venue and year. This costs 0.10 (Sonnet~4.6) and 0.12 (GPT-5.4) core, almost entirely on \texttt{sc:datePublished} and \texttt{cr:citeAs} (grey rows). Docling's default parser mishandles one paper, a 2009 technical report set in legacy Type~3 bitmap fonts: it resolves their glyph names as ZapfDingbats symbols, producing letter-substituted text (its pdfium backend reads the paper correctly). We keep the default parser, since the ablation tests Docling as it ships.

\begin{table}[!ht]
    \centering
    \caption{\textbf{PDF parser ablation.} Same prompt and scoring pipeline; only the text extractor varies. 30 test-split papers (seed 42). Docling uses the default converter with the full export (body, page headers/footers, figure text); grey rows use Docling's default body-only export. RAI cells for all rows were judged in one session, so compare within this table only. $^\dagger$29 papers: Docling's default parser resolves one paper's legacy Type~3 glyph names as ZapfDingbats symbols (its pdfium backend reads it correctly); Sonnet~4.6 refuses the resulting text.}
    \label{tab:parser_ablation}
    \small
    \begin{tabular}{llccc}
    \toprule
    Model & Parser & Core & RAI & Overall \\
    \midrule
    \multirow{4}{*}{Sonnet~4.6}
     & PyPDF2 (ours)                    & \textbf{0.770} & 0.475 & \textbf{0.573} \\
     & PyMuPDF                          & 0.748 & 0.475 & 0.566 \\
     & Docling, full export$^\dagger$   & 0.749 & 0.472 & 0.564 \\
     & \textcolor{gray}{Docling, body only$^\dagger$} & \textcolor{gray}{0.670} & \textcolor{gray}{0.478} & \textcolor{gray}{0.542} \\
    \midrule
    \multirow{4}{*}{GPT-5.4}
     & PyPDF2 (ours)                    & \textbf{0.670} & 0.464 & 0.532 \\
     & PyMuPDF                          & \textbf{0.670} & 0.473 & \textbf{0.541} \\
     & Docling, full export             & 0.656 & 0.477 & 0.539 \\
     & \textcolor{gray}{Docling, body only} & \textcolor{gray}{0.548} & \textcolor{gray}{0.507} & \textcolor{gray}{0.521} \\
    \bottomrule
    \end{tabular}
\end{table}

\subsection{Architectural Failure-Mode Analysis}
\label{app:arch-gap}

This appendix expands on the error analysis in Section~\ref{sec:error-analysis}. For each agentic architecture using Sonnet~4.6, we randomly sampled RAI cells where the agentic score was lower than the single-pass Sonnet~4.6 score for the same paper and field. The sample contains 30 cells per architecture, except Triage~+~Critique, which has 27, with at most two cells per field for each architecture. We compared the reference answer, both system outputs, and the paper text, and assigned one error type to each cell. An LLM assisted with the initial labels, which an author reviewed and confirmed. Consistent with the benchmark, we treated the reference answer as authoritative. Substantive discrepancies from the reference were classified as system errors; answers that matched the reference in substance but received a lower score were classified as \emph{judge noise}.

\paragraph{Failure-mode Counts by Architecture}

Table~\ref{tab:failure-modes-by-arch} summarizes the error counts in the sampled cells. Leaving documented fields empty is the most common error for ReAct, Parallel Specialists, and Triage~+~Critique; incomplete answers are the most common error for Locator-Extractor. Some empty fields occur even when the system reports the relevant information elsewhere. In 5 of the 13 empty Parallel Specialists cells, another specialist includes the information under a different field. In 8 of the 15 empty ReAct cells, the agent includes it under another field or mentions it in its explanation for leaving the target field empty. Judge noise accounts for 6 of the 117 cells. The exhibits below show the reference answer alongside the single-pass and agentic outputs.

\begin{table}[!ht]
    \centering
    \footnotesize
    \setlength{\tabcolsep}{4pt}
    \caption{Error types in 117 sampled RAI cells where an agentic system using Sonnet~4.6 scored below single-pass Sonnet~4.6 on the same paper and field. Each cell receives one label, with the reference answer treated as authoritative. Percentages are calculated within each column and rounded to the nearest integer.}
    \label{tab:failure-modes-by-arch}
    \begin{tabular}{lccccc}
    \toprule
    Error type & \makecell{ReAct\\(n=30)} & \makecell{Parallel Specialists\\(n=30)} & \makecell{Triage + Critique\\(n=27)} & \makecell{Locator-Extractor\\(n=30)} & \makecell{Total\\(n=117)} \\
    \midrule
    left empty although documented & 15 (50\%) & 13 (43\%) & 13 (48\%) & 10 (33\%) & 51 (44\%) \\
    incomplete & 8 (27\%) & 10 (33\%) & 9 (33\%) & 13 (43\%) & 40 (34\%) \\
    wrong detail & 5 (17\%) & 5 (17\%) & 3 (11\%) & 5 (17\%) & 18 (15\%) \\
    judge noise & 2 (7\%) & 2 (7\%) & 2 (7\%) & 0 & 6 (5\%) \\
    \makecell[l]{other (verbose drift,\\ over-hedged paraphrase)} & 0 & 0 & 0 & 2 (7\%) & 2 (2\%) \\
    \bottomrule
    \end{tabular}
\end{table}

\paragraph{Parallel Specialists Exhibits}
The most common error in the Parallel Specialists sample is leaving a documented field empty (13 of 30 cells). In 5 of these cells, another specialist reports the relevant information under a different field.

\textbf{Exhibit SPECIALIST-1.} \texttt{MMLU\_\allowbreak 30field} / \texttt{rai:annotatorDemographics}. \emph{Error type:} left empty although documented.
\begin{description}[leftmargin=4em,style=nextline,itemsep=0pt,topsep=2pt]
  \item[Gold:] Graduate and undergraduate students
  \item[Single-Pass:] Graduate and undergraduate students
  \item[Parallel Specialists:] (empty)
  \item[What happened:] The responsible specialist leaves this field empty, while other specialists report ``graduate and undergraduate students'' under data collection and annotation protocol.
\end{description}

\textbf{Exhibit SPECIALIST-2.} \texttt{FLORES\_\allowbreak 30field} / \texttt{rai:dataCollectionType}. \emph{Error type:} incomplete.
\begin{description}[leftmargin=4em,style=nextline,itemsep=0pt,topsep=2pt]
  \item[Gold:] Manual Human Curator, Document analysis, Web Scraping
  \item[Single-Pass:] Manual Human Curator, Document analysis, Web Scraping
  \item[Parallel Specialists:] Manual Human Curator; Document analysis
  \item[What happened:] The answer includes two of the three reference categories but omits Web Scraping.
\end{description}

\paragraph{Triage + Critique Exhibits}
The most common error in the Triage~+~Critique sample is leaving a documented field empty (13 of 27 cells).

\textbf{Exhibit V2-1.} \texttt{Idavidrein\_\allowbreak gpqa} / \texttt{rai:dataAnnotationPlatform}. \emph{Error type:} left empty although documented.
\begin{description}[leftmargin=4em,style=nextline,itemsep=0pt,topsep=2pt]
  \item[Gold:] Custom data collection platform built for this project, with contractors recruited and managed through Upwork.
  \item[Single-Pass:] Upwork (freelancer platform for contractor recruitment and management); a custom data collection platform built by the authors.
  \item[Triage + Critique:] (empty)
  \item[What happened:] The system leaves the field empty although the paper identifies both Upwork and the authors' custom collection platform.
\end{description}

\textbf{Exhibit V2-2.} \texttt{livecodebench\_\allowbreak code\_\allowbreak generation\_\allowbreak lite} / \texttt{rai:dataCollectionType}. \emph{Error type:} incomplete.
\begin{description}[leftmargin=4em,style=nextline,itemsep=0pt,topsep=2pt]
  \item[Gold:] Web Scraping, Software Collection
  \item[Single-Pass:] Web Scraping, Software Collection
  \item[Triage + Critique:] Web Scraping
  \item[What happened:] The answer includes Web Scraping but omits the second reference category, Software Collection.
\end{description}

\paragraph{Locator-Extractor Exhibits}
Incomplete answers are the most common error in the Locator-Extractor sample (13 of 30 cells).

\textbf{Exhibit LEV-1.} \texttt{mrmrx\_\allowbreak CADS-dataset} / \texttt{rai:dataCollectionTimeframe}. \emph{Error type:} left empty although documented.
\begin{description}[leftmargin=4em,style=nextline,itemsep=0pt,topsep=2pt]
  \item[Gold:] 2007-2024
  \item[Single-Pass:] Start: 2007, End: 2024
  \item[Locator-Extractor:] (empty)
  \item[What happened:] The system leaves the field empty although the main dataset description states the collection period.
\end{description}

\textbf{Exhibit LEV-2.} \texttt{Rowan\_\allowbreak hellaswag} / \texttt{rai:dataCollectionType}. \emph{Error type:} incomplete.
\begin{description}[leftmargin=4em,style=nextline,itemsep=0pt,topsep=2pt]
  \item[Gold:] Web Scraping, Software Collection, Manual Human Curator
  \item[Single-Pass:] Web Scraping, Software Collection, Manual Human Curator
  \item[Locator-Extractor:] Web Scraping
  \item[What happened:] The answer includes only Web Scraping. Its supporting quote identifies ActivityNet and WikiHow as sources but does not mention the other collection steps.
\end{description}

\paragraph{ReAct Exhibits}
The most common error in the ReAct sample is leaving a documented field empty (15 of 30 cells). In 8 of these cells, the agent reports the relevant information under another field or mentions it in its explanation for leaving the target field empty.

\textbf{Exhibit REACT-1.} \texttt{rajpurkar\_\allowbreak squad} / \texttt{rai:dataCollectionMissingData}. \emph{Error type:} left empty although documented.
\begin{description}[leftmargin=4em,style=nextline,itemsep=0pt,topsep=2pt]
  \item[Gold:] Over the development and test sets, 2.6\% of questions were marked unanswerable by at least one of the additional crowdworkers.
  \item[Single-Pass:] Over the development and test sets, 2.6\% of questions were marked unanswerable by at least one of the additional crowdworkers during the secondary answer generation task.
  \item[ReAct:] (empty)
  \item[What happened:] The agent reports the 2.6\% figure for unanswerable questions under annotation analysis but leaves this field empty.
\end{description}

\textbf{Exhibit REACT-2.} \texttt{NLPCoreTeam\_\allowbreak mmlu\_\allowbreak ru} / \texttt{rai:annotatorDemographics}. \emph{Error type:} left empty although documented.
\begin{description}[leftmargin=4em,style=nextline,itemsep=0pt,topsep=2pt]
  \item[Gold:] Graduate and undergraduate students
  \item[Single-Pass:] Graduate and undergraduate students collected the questions. Unspecialized humans from Amazon Mechanical Turk were used to establish a human-level accuracy baseline, obtaining 34.5\% accuracy on the test.
  \item[ReAct:] (empty)
  \item[What happened:] The agent leaves the field empty even though its explanation states that graduate and undergraduate students collected the questions.
\end{description}

\subsection{Popularity, Recency and Documentation Sparsity}
\label{app:difficulty}

Two natural concerns are that systems score well because they have seen popular or older datasets during pretraining, and that long papers are simply harder to read. Neither explains the scores on the test split. We collected Hugging Face download counts for 87 of the 88 test datasets, which span three orders of magnitude from 21.9 thousand to 42.5 million downloads, together with publication dates from the arXiv identifiers. For all major systems the correlation between score and download count is close to zero, for example Spearman $\rho = 0.07$ for Sonnet~4.6 and $-0.11$ for GPT-5.4, and no correlation survives FDR correction. Recent papers are also extracted as accurately as older ones: Sonnet~4.6 scores 0.754 on the 26 papers from 2024 and 2025 and 0.756 on papers from 2020 or earlier. Within the popularity range the benchmark covers, we therefore find no evidence that familiarity rather than reading drives the scores, although these analyses cannot rule out pretraining contamination entirely.

Paper length has no measurable effect either: for Sonnet~4.6, the correlation between per-paper score and page count is $-0.01$ across papers of 6 to 119 pages, and mean scores are stable across length quartiles. What matters is how much a paper documents. The share of a paper's cells with no documented value correlates with its score at $\rho = -0.40$ for Sonnet~4.6 and $-0.61$ for GPT-5.4, and scores fall steadily from the best-documented to the most sparsely documented quartile, from 0.789 to 0.694 for Sonnet~4.6 and from 0.748 to 0.620 for GPT-5.4. What makes a paper hard is therefore not reading a longer document but deciding correctly what is absent.

\subsection{Scaling Within a Model Family}
\label{app:qwen-sweep}

Comparing models from different families mixes model size with differences in training data and design. To isolate size, we ran four Qwen3 models with 4B, 8B, 14B and 32B parameters as single-pass extractors under the same protocol, on the same test papers and with the full paper text. One test paper exceeds the context window and is excluded for all four sizes, and on the 66 papers where all four produced valid output every size is scored on exactly the same cells; Table~\ref{tab:qwen-sweep} lists the results. Composite scores rise steadily with size, from 0.376 for Qwen3-4B to 0.528 for Qwen3-32B. The gain comes almost entirely from the long-form RAI fields, which nearly double from 0.272 to 0.484, while the constrained core fields stay within a narrow band. Even the 32B model remains well below the frontier single-pass systems in Table~\ref{tab:test88-headline}, so scaling within an open-weight family does not close that gap.

\begin{table}[!ht]
    \centering
    \footnotesize
    \caption{Qwen3 models of increasing size as single-pass extractors, scored on the same cells of 66 test papers.}
    \label{tab:qwen-sweep}
    \begin{tabular}{lccc}
    \toprule
    Model & Composite [95\% CI] & Core & RAI \\
    \midrule
    Qwen3-4B  & 0.376 [0.359, 0.394] & 0.586 & 0.272 \\
    Qwen3-8B  & 0.431 [0.411, 0.452] & 0.577 & 0.358 \\
    Qwen3-14B & 0.482 [0.458, 0.506] & 0.659 & 0.393 \\
    Qwen3-32B & 0.528 [0.505, 0.554] & 0.615 & 0.484 \\
    \bottomrule
    \end{tabular}
\end{table}

\section{Per-System Cost}
\label{app:cost}

Table~\ref{tab:cost-detail} reports mean per-paper inference costs for the systems in Table~\ref{tab:test88-headline}, estimated from recorded token usage at standard public list prices. We apply the same pricing basis to single-pass and agentic runs, including single-pass runs executed through a discounted batch API. The estimates exclude papers with unrecorded token usage and account for caching only where cached-token counts are available; unrecorded Gemini thinking tokens are omitted.

\begin{table}[!ht]
    \centering
    \small
    \setlength{\tabcolsep}{4pt}
    \caption{Estimated mean per-paper inference costs for the systems in Table~\ref{tab:test88-headline}, based on recorded token usage on the 88-paper test split and standard public list prices at the time of the experiments. Means exclude papers with unrecorded token usage. Ratios compare each system with single-pass extraction on the same backbone; the hybrid is compared with single-pass Gemini~3.1~Pro. $^{\dag}$Self-hosted on an institutional cluster with no API charge; compute costs are not included. $^{\ddag}$Unrecorded Gemini thinking tokens are excluded, understating output-token costs. Inputs are priced as uncached where cached-token counts were not recorded. $^{\S}$ReAct sends a growing prompt at each turn. If all prompt tokens after the first turn were cache hits, the estimated costs would be \$0.146 for GPT-5.4 and \$0.103 for Gemini~3.1~Pro, still excluding Gemini thinking tokens.}
    \label{tab:cost-detail}
    \begin{tabular}{llrr}
    \toprule
    System & Architecture & USD per paper & vs.\ single-pass \\
    \midrule
    Claude Sonnet 4.6 & Single-Pass & 0.126 & \\
    Claude Opus 4.7 & Single-Pass & 0.269 & \\
    GPT-5.4 & Single-Pass & 0.093 & \\
    Qwen 3.6 35B-A3B & Single-Pass & 0$^{\dag}$ & \\
    GLM-5.1 & Single-Pass & 0.063 & \\
    Gemini 2.5 Flash & Single-Pass & 0.013$^{\ddag}$ & \\
    GPT-5.4 Mini & Single-Pass & 0.026 & \\
    Gemini 3.1 Pro Preview & Single-Pass & 0.073$^{\ddag}$ & \\
    DeepSeek V3.2 & Single-Pass & 0.008 & \\
    Mistral Small 4 & Single-Pass & 0$^{\dag}$ & \\
    Llama 4 Scout 17B & Single-Pass & 0$^{\dag}$ & \\
    \midrule
    ReAct (Sonnet 4.6) & ReAct & 0.261 & 2.1$\times$ \\
    ReAct (GPT-5.4) & ReAct & 0.542$^{\S}$ & 5.8$\times$ \\
    ReAct (Gemini 3.1 Pro) & ReAct & 0.304$^{\ddag\S}$ & 4.2$\times$ \\
    \midrule
    Parallel Specialists (Sonnet 4.6) & Parallel Specialists & 0.726 & 5.7$\times$ \\
    Parallel Specialists (GPT-5.4) & Parallel Specialists & 0.531 & 5.7$\times$ \\
    Parallel Specialists (Gemini 3.1 Pro) & Parallel Specialists & 0.568$^{\ddag}$ & 7.8$\times$ \\
    \midrule
    Triage + Critique (Sonnet 4.6) & Triage + Critique & 0.213 & 1.7$\times$ \\
    Triage + Critique (GPT-5.4) & Triage + Critique & 0.162 & 1.7$\times$ \\
    Triage + Critique (Gemini 3.1 Pro) & Triage + Critique & 0.136$^{\ddag}$ & 1.9$\times$ \\
    \midrule
    Locator-Extractor (Sonnet 4.6) & Locator-Extractor & 0.195 & 1.5$\times$ \\
    Locator-Extractor (GPT-5.4) & Locator-Extractor & 0.110 & 1.2$\times$ \\
    Locator-Extractor (Gemini 3.1 Pro + GPT-5.4 Mini) & Locator-Extractor & 0.076$^{\ddag}$ & 1.0$\times$ \\
    Locator-Extractor (Gemini 3.1 Pro) & Locator-Extractor & 0.091$^{\ddag}$ & 1.2$\times$ \\
    \midrule
    \multicolumn{4}{l}{\textit{Unranked diagnostic}} \\
    Claude Sonnet 4.5 & Single-Pass & 0.121 & \\
    \bottomrule
    \end{tabular}
\end{table}

\section{Agentic Configurations: Full Backbone Sweep}
\label{app:agentic-full}

The headline architectural-isolation analysis (Section~\ref{sec:headline-ranking}) pins all four agentic patterns to the Sonnet~4.6 backbone for the cleanest comparison. We also evaluated cross-backbone variants where compute and time allowed; their composites appear in Table~\ref{tab:test88-headline}, except for the Sonnet~4.5 variants, which are excluded from the headline table as gold-seed runs, and are summarised here for reference.

\textbf{Triage~+~Critique Sweep.} Tested on Sonnet~4.5 (composite 0.667; gold-seed and excluded from the headline ranking), Sonnet~4.6 (0.624), GPT-5.4 (0.557), and Gemini~3.1~Pro (0.436). The within-backbone direction (Triage~+~Critique below the matching single-pass) holds in every case.

\textbf{Locator-Extractor Sweep.} Tested on Sonnet~4.5 (0.556; gold-seed), Sonnet~4.6 (0.566), GPT-5.4 (0.502), Gemini~3.1~Pro (0.448), and the Gemini~3.1~Pro~+~GPT-5.4-Mini hybrid (0.478). Locator-Extractor is the lowest agentic configuration on Sonnet~4.6 and GPT-5.4 and the second lowest on Gemini~3.1~Pro, just above Triage~+~Critique (0.436), matching the per-architecture failure-mode analysis (Section~\ref{sec:error-analysis}, Appendix~\ref{app:arch-gap}): the locator-buffer approach is the most decomposed of the four patterns and bears the largest accuracy loss.

\textbf{Parallel Specialists Sweep.} Tested on Sonnet~4.6 (0.647), GPT-5.4 (0.589), and Gemini~3.1~Pro (0.539). The within-backbone direction (Parallel Specialists below the matching single-pass) holds on every backbone.

\textbf{ReAct Sweep.} Tested on Sonnet~4.6 (0.652), GPT-5.4 (0.631), and Gemini~3.1~Pro (0.582); the within-backbone direction holds on every backbone. Earlier iterations of ReAct on Sonnet~4.6 (variants \texttt{[v1]} and \texttt{[iter-5]}) were run but are excluded from the headline ranking to avoid double-counting; the \texttt{[v3]} configuration is the released variant.

\section{Limitations and Future Work}
\label{app:limitations}

This appendix expands on the limitations of the \cm benchmark summarized in Section~\ref{sec:limitations} and discusses additional methodological caveats that shape how the benchmark and its results should be interpreted.

\textbf{PDF Parser Noise.} Section~\ref{sec:task-pipeline} extracts text with PyPDF2, which linearises tables and figure text and leaves equations unparsed. We tested whether a better parser helps by re-running extraction with PyMuPDF and with the layout-aware Docling (Appendix~\ref{app:parser-ablation}). With Docling's full export (body text, page headers/footers, and text inside figures), overall scores stay within 0.01 of PyPDF2 for both models (Sonnet~4.6: 0.564 vs.\ 0.573; GPT-5.4: 0.539 vs.\ 0.532), and there is no consistent gain on the table-heavy fields \texttt{rai:annotationsPerItem} and \texttt{rai:dataAnnotationAnalysis}. Errors on these fields therefore appear to stem from how the evidence is reported and interpreted rather than from text extraction. Vision-language preprocessing, which reads tables and figures as images, remains untested.

\textbf{English-only Gold Split.} The 102 gold papers were chosen by Hugging Face download rank, which heavily favours English-language datasets. The Croissant schema fields themselves are language-agnostic, but reporting conventions differ substantially across language communities (e.g., where licensing information is typically stated, how annotator demographics are described). Multilingual gold annotation with prompt localization for non-English papers is the natural extension; it would require recruiting annotators with the relevant language coverage, which we defer to future work.

\textbf{Benchmark Scale and Contamination Caveat.} The 102-paper gold split supports per-field accuracy estimates with 2{,}000-replicate paper-clustered bootstrap CIs but is too small for tight per-domain stratification (vision, NLP, multimodal, audio, code, robotics, medical). The 500-paper silver split addresses scale for distillation and corpus-trend studies but cannot replace gold for evaluation. Separately, we cannot rule out training-data contamination: most of the 102 papers pre-date the model release windows of the systems we evaluate, and we have no model-side training-data disclosure that would let us run a contamination test along the lines of~\citep{oren2023proving}.

\textbf{Gold-seed Dependency.} Gold values were derived from human verification of Claude Sonnet 4.5 pre-fills, then corrected through an end-to-end author audit that revised 191 of 3{,}060 cells, spread across 78 of the 102 papers and 25 of the 30 fields. The audit catches the substantive errors (mis-extractions, hallucinations, granularity mismatches) but the residual structural pattern of what the seed model marks as ``filled'' versus ``null'' is inherited by the gold. This is why Sonnet~4.5 is excluded from the headline ranking in Table~2: its composite would be inflated by construction. Appendix~\ref{app:seed-robustness} tests this directly: re-annotating 10 papers from GPT-5.4 pre-fills leaves the ranking of the other systems stable, while the seed model and, to a lesser degree, its family gain.

\textbf{Single LLM Judge.} The Tier~2 judge is a single model (GLM-5) selected from six candidates via a pre-registered protocol on a 60-cell calibration + validation split (Section~\ref{sec:experimental-setup}, Table~\ref{tab:judge-selection}). The pooled agreement of $\kappa = 0.890$ refers to judge selection on 60 development cells using the initial rubric and pre-audit gold labels. With the final rubric and audit-corrected gold, the deployed judge achieves $\kappa = 0.708$ on the same cells and $\kappa = 0.663$ on the separate 200-cell audit. We did not run a multi-judge ensemble for the headline composite because the protocol's tiebreak rule (locked 2026-04-27) selects a single judge, and ensembling re-introduces the calibration question across judges. Future work could evaluate whether a multi-judge ensemble improves agreement with human judgments and assess the sensitivity of system rankings to judge selection.

\textbf{Residual LLM Non-determinism.} All extractions fix temperature to 0, but frontier models over long-context inputs are not strictly deterministic. The bootstrap intervals capture paper-level sampling uncertainty, not run-to-run model stochasticity. We mitigate by pinning model versions in the \texttt{\_meta} block and stamping the prompt SHA-256 prefix into every extraction, but the field-level scoring variance from a re-run of the same model on the same paper is not separately quantified.

\textbf{Architecture Coverage.} The headline architectural-isolation analysis (Section~\ref{sec:headline-ranking}, Section~\ref{sec:error-analysis}) pins all four agentic patterns to the Sonnet~4.6 backbone. Cross-backbone variants of all four architectures (on GPT-5.4 and Gemini~3.1~Pro, plus the Gemini-3.1-Pro~+~GPT-5.4-Mini locator hybrid) are reported in Table~2 and confirm the within-backbone direction: the matching single-pass configuration outperforms the agentic variant in all twelve architecture-backbone cells.

\textbf{Field-category Scope.} The 30-field Croissant 1.1 schema is the largest schema-aligned benchmark we are aware of for ML dataset metadata, but it is not exhaustive: structural fields (column types, distribution statistics) that platforms like Hugging Face auto-generate are out of scope, as are multi-resource datasets where individual sub-resources have separate metadata. Extension to Croissant~2.x or to a richer schema with sub-resource fields is straightforward and would multiply the per-paper extraction work proportionally.

\end{document}